\documentclass[10pt,twocolumn,letterpaper]{article}

\usepackage[pagenumbers]{cvpr} 

\usepackage[dvipsnames]{xcolor}
\newcommand{\myfirstpara}[1]{\par \noindent \textbf{{#1}:}}
\newcommand{\mypara}[1]{\vspace{0.2em}\myfirstpara{#1}}
\newcommand{\cmark}{{\color{green}\ding{51}}}
\newcommand{\xmark}{{\color{red}\ding{55}}}
\usepackage{comment}
\usepackage{amssymb,comment,enumitem}
\usepackage{graphicx,amsmath,bm}
\usepackage{enumitem}

\def\lomo{\texttt{LoMOE}\xspace}

\def\proposedDataset{\texttt{LoMOE}-Bench\xspace}
\definecolor{cvprblue}{rgb}{0.21,0.49,0.74}
\usepackage[pagebackref,breaklinks,colorlinks,citecolor=cvprblue]{hyperref}
\usepackage{amssymb}
\usepackage{pifont}
\usepackage{makecell}
\usepackage{adjustbox}

\usepackage{algorithm}
\usepackage{algorithmic}

\def\paperID{12811} 
\def\confName{CVPR}
\def\confYear{2024}

\title{LoMOE: Localized Multi-Object Editing via Multi-Diffusion}

\author{Goirik Chakrabarty$^{\,1 *}$ \; Aditya Chandrasekar$^{\,2 *}$ \; Ramya Hebbalaguppe$^{\,1, 3}$ \; Prathosh AP$^{\,2}$\\
$^{1}\,$TCS Research \quad $^{2}\,$IISc Bangalore \quad $^{3}\,$IIT Delhi\\
}

\begin{document}


\twocolumn[{%
	\renewcommand\twocolumn[1][]{#1}%
	\maketitle
	\begin{center}
		\vspace{-1em}
    \includegraphics[width=\textwidth]{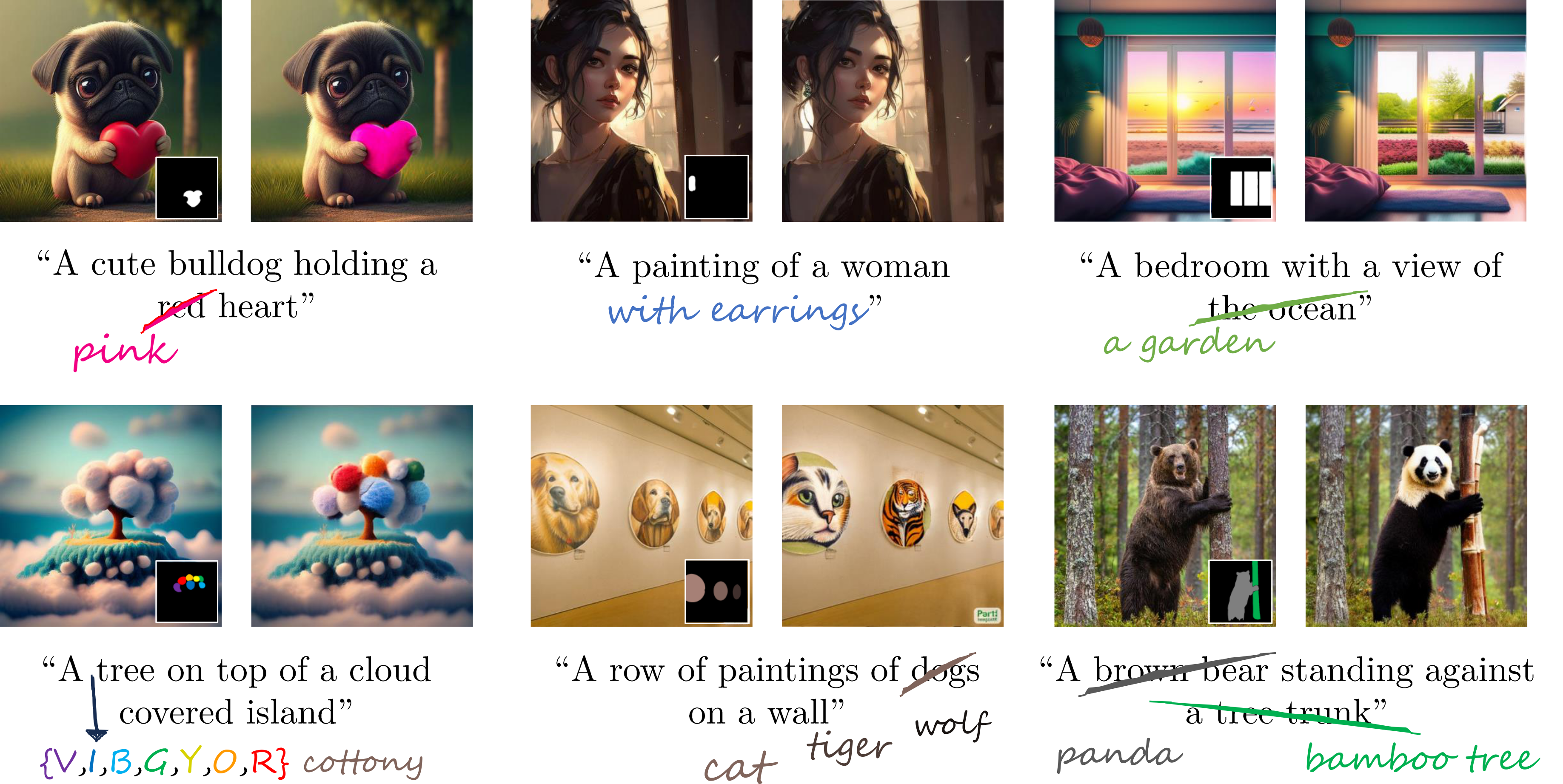}
   \captionof{figure}{\textbf{Representative results of \lomo on diverse images}: Our algorithm can handle both single and multi-object edits in one go. The first image in each example depicts the original image with the input mask (specifying the edit locations). Below each image is the original text used for its generation and the input text prompt (colored font) describing the edits. The second image depicts the edited image via our method. It is seen that our method can handle intricate localized object details such as heart color, earrings, window-view, multiple-cloud coloring, animal types in a painting, and tree-animal type.}
		\label{fig:teaser}
	\end{center}%
}]

\begingroup\renewcommand\thefootnote{\textsection}
\footnotetext{$^*$ denotes equal contribution.}
\endgroup
\addtocontents{toc}{\protect\setcounter{tocdepth}{-2}}
\begin{abstract}
Recent developments in the field of diffusion models have demonstrated an exceptional capacity to generate high-quality prompt-conditioned image edits.  
Nevertheless, previous approaches have primarily relied on textual prompts for image editing, which tend to be less effective when making precise edits to specific objects or fine-grained regions within a scene containing single/multiple objects. We introduce a novel framework for zero-shot localized multi-object editing through a multi-diffusion process to overcome this challenge. This framework empowers users to perform various operations on objects within an image, such as adding, replacing, or editing \textbf{many} objects in a complex scene \textbf{in one pass}. Our approach leverages foreground masks and corresponding simple text prompts that exert localized influences on the target regions resulting in high-fidelity image editing. A combination of cross-attention and background preservation losses within the latent space ensures that the characteristics of the object being edited are preserved while simultaneously achieving a high-quality, seamless reconstruction of the background with fewer artifacts compared to the current methods. We also curate and release a dataset dedicated to multi-object editing, named \proposedDataset. Our experiments against existing state-of-the-art methods demonstrate the improved effectiveness of our approach in terms of both image editing quality and inference speed.


\end{abstract}

\section{Introduction}
\label{sec:intro}

Diffusion models \cite{dall-e, stable-diff, ddim} have exhibited an outstanding ability to generate highly realistic images based on text prompts. 
However, text-based editing of multiple fine-grained objects precisely at given locations within an image is a challenging task. This challenge primarily stems from the inherent complexity of controlling diffusion models to specify the accurate spatial attributes of an image, such as the scale and occlusion during synthesis. 
Existing methods for textual image editing use a global prompt for editing images, making it difficult to edit in a specific region while leaving other regions unaffected~\cite{sdedit,instruct-p2p}. Thus, this is an important problem to tackle, as real-life images often have multiple subjects and it is desirable to edit each subject independent of other subjects and the background while still retaining coherence in the composition of the image. This forms the objective of our work, called \textbf{Lo}calized \textbf{M}ulti-\textbf{O}bject \textbf{E}diting (\lomo).


Our method draws inspiration from the recent literature on compositional generative models~\cite{multidiff, dense-t2i-compose, localized-compose}. 
It inherits generality without requiring training, making it a zero-shot solution similar to \cite{multidiff}. We utilize a pre-trained StableDiffusion 2.0~\cite{stable-diff} as our base generative model. Our approach involves the manipulation of the diffusion trajectory within specific regions of an image earmarked for editing. We employ prompts that exert a localized influence on these regions while simultaneously incorporating a global prompt to guide the overall image reconstruction process that ensures a coherent composition of foreground and background with minimal/imperceptible artifacts. To initiate our editing procedure, we employ the inversion of the original image as a starting point, as proposed in \cite{zero-p2p}. For achieving high-fidelity, human-like edits in our images, we employ two crucial steps: \textbf{(a)} cross-attention matching and \textbf{(b)} background preservation. These preserve the integrity of the edited image by guaranteeing that the edits are realistic and aligned with the original image. This, in turn, enhances the overall quality and perceptual authenticity of the final output. Additionally, we also curate a novel benchmark dataset,  named \proposedDataset for multi-object editing. Our contributions in this paper are as follows:

\begin{enumerate}
    \item We present a framework called \lomo for zero-shot text-based localized multi-object editing based on Multi-diffusion \cite{multidiff}.
    \item Our framework facilitates multiple edits in a single iteration via enforcement of cross-attention and background preservation, resulting in high fidelity and coherent image generation.
    \item We introduce a new benchmark dataset for evaluating the multi-object editing performance of existing frameworks, termed \proposedDataset. 
\end{enumerate}


\section{Related Work}

\subsection{Image Synthesis and Textual Guidance}

Text-to-image synthesis has made significant strides in recent years, with its early developments rooted in RNNs~\cite{rnn} and GANs~\cite{goodfellow2014generative}, which were effective in generating simple objects such as flowers, dogs and cats but struggled in generating complex scenes, especially with multiple objects~\cite{gan-cannot}. These models have now been superseded by diffusion-based methods which produce photorealistic images, causing a paradigm shift~\cite{ddpm, ddim, stable-diff}.

In a separate line of work, CLIP~\cite{clip} was introduced, which is a vision-language model trained on a dataset of 400 million image-text pairs using techniques such as contrastive training. The rich embedding space CLIP provides has enabled various multi-modal applications such as text-based imaged generation~\cite{vqgan-clip, wang2022clip-gen, clip-gen2, gal2022stylegan, dhariwal2021diffusion, stable-diff, imagic, dall-e}.

\subsection{Compositional Diffusion Model}

As observed by Kim \etal \cite{dense-t2i-compose}, text-to-image models fail to adhere to the positional/layout prompting via text. Therefore, compositional diffusion models try to address the task of image generation conditioned on masks, where each mask is associated with a text prompt. In Make-a-Scene~\cite{make-a-scene}, the initial step involves predicting a segmentation mask based on the provided text. Subsequently, this generated mask is employed in conjunction with the text to produce the final predicted image. Methods such as Controlnet and Gligen \cite{controlnet, gligen} have proposed fine-tuning for synthesizing images given text descriptions and spatial controls based on adapters. Finally, methods like ~\cite{multidiff, localized-compose, dense-t2i-compose}, try to utilise the pre-trained models and masked regions with independent prompts to generate images without re-training.

\subsection{Image Editing}

Paint-by-Word~\cite{paint-by-word} was one of the first approaches to tackle the challenge of zero-shot local text-guided image manipulation. 
But this method exclusively worked with generated images as input and it required a distinct generative model for each input domain. Later, Meng \etal \cite{sdedit} showed how the forward diffusion process allows image editing by finding a common starting point for the original and the editing image. This popularised inversion among image editing frameworks such as ~\cite{imagic, zero-p2p}. This approach was further improved upon by adding a structure prior to the editing process using cross-attention matching~\cite{zero-p2p, prompt2prompt}. Moreover, there have been improvements in inversion techniques producing higher quality reconstruction which results in more faithful edits~\cite{null-text, direct-inversion}.
However, many of the aforementioned methods generate the whole image from the inversion. This compromises the quality of reconstruction in regions where the image was not supposed to be edited. 

For reliable editing, it is essential that the generation process is restricted to a certain localized region~\cite{bld, diffedit, glide}. These methods fall short on two counts: (1) editing multiple regions in one pass, and (2) maintaining consistency between the edited and the non-edited regions of the image. Our method explicitly takes care of these two aspects of image editing, while incorporating all the advancements of our predecessor methods.
\section{Proposed Method}

\mypara{Problem Statement} In a multi-object editing scenario, the objective is to simultaneously make local edits to several objects within an image. Formally, we are given a pretrained diffusion model $\Phi$, an image $\mathbf{x}_0 \in \mathcal{X}$, and $N$ binary masks $\{M_1, \cdots, M_N\}$ along with a corresponding set of prompts $\{c_1, \cdots, c_N\}$, where $c_i \in \mathcal{C}$, the space of encoded text prompts. These are used to obtain an edited image $\mathbf{x}^{*}$ such that the editing process precisely manifests at the locations dictated by the masks, in accordance with the guidance provided by the prompts. 

\begin{figure}[t]
    \adjustbox{max width=0.9\linewidth}{\centering
\includegraphics[width=\textwidth]{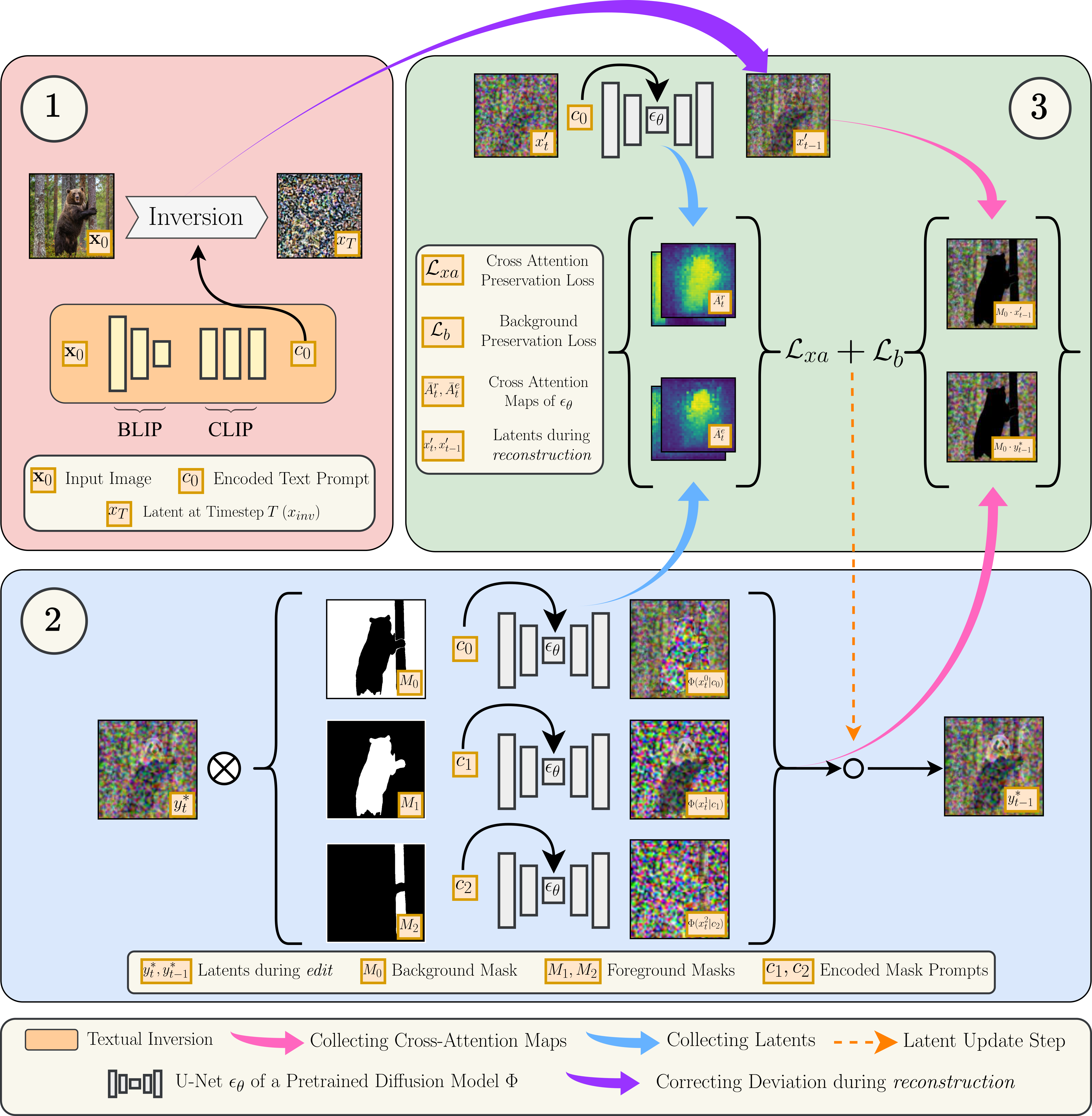} }
    \caption{\textbf{Overview of the proposed LoMOE framework}: \ding{192} Inversion, to obtain $x_{inv}$ and $c_0$ corresponding to the input image $\mathbf{x}_0$. \ding{193} MultiDiffusion process to restrict the edits to masked regions $M_1, M_2$ guided by $c_1, c_2$. \ding{194} Preservation of Attributes, via $\mathcal{L}_{xa}$ and $\mathcal{L}_b$ using reference cross-attention maps and background latents using a \textit{reconstruction} process.}
    \label{fig:pipeline}
\end{figure}

\mypara{Overview of LoMOE} Our proposed method {\bf Lo}calized {\bf M}ulti-{\bf O}bject image {\bf E}diting (\lomo)  
comprises of three key steps \textbf{(a)} Inversion of the original image $\mathbf{x}_0$ to obtain the latent code $x_{inv}$, which initiates the editing procedure and ensures a coherent and controlled edit \textbf{(b)} Applying the MultiDiffusion process for localized multi-object editing to limit the edits to mask-specific regions, and \textbf{(c)} Attribute and Background Preservation via cross attention and latent background preservation to retain structural consistency with the original image. Figure \ref{fig:pipeline} depicts the overview of our method. 

\subsection{Inversion for Editing}

In this work, we employ a pretrained Stable Diffusion~\cite{stable-diff} model, denoted as $\Phi$. This model encodes an input image $\mathbf{x}_0 \in \mathcal{X}=\mathbb{R}^{512 \times 512 \times 3}$ into a latent code $x_0 \in \mathbb{R}^{64 \times 64 \times 4}$.

Given an image $\mathbf{x}_0$ and it's corresponding latent code $x_0$, \textit{inversion} entails finding a latent $x_{inv}$ which reconstructs $x_0$ upon sampling. In a DDPM \cite{ddpm}, the \textit{inversion} step is defined by the forward diffusion process, which involves Gaussian noise perturbation $\big(\epsilon_t \sim \mathcal{N}(0,\,I)\big)$  for a fixed number of timesteps $t \in [T]$, governed by \cref{eq:1,eq:2}.
\begin{gather}
    \label{eq:1}
    x_t = \sqrt{\alpha_t} x_0 + \sqrt{1 - \alpha_t} \epsilon_t\\
    x_{inv} = x_T    
    \label{eq:2}
\end{gather} where $\alpha_t$ represents a prefixed noise schedule. But given the stochastic nature of the DDPM forward and reverse process which leads to poor reconstruction upon sampling, we adopt a deterministic DDIM reverse process~\cite{ddim, zero-p2p} which is deterministic when $\sigma_t = 0$ $\forall$ $t$, where the family $\mathcal{Q}$ of inference distributions is parameterized by $\sigma \in \mathbb{R}^{T}_{+}$.
\begin{multline}
x_{t+1}=\sqrt{\alpha_{t+1}} \left(\frac{x_t - \sqrt{1-\alpha_{t}} \: \epsilon_\theta\left(x_t, t\right)}{\sqrt{\alpha_t}}\right)\\+ \sqrt{1-\alpha_{t+1}} \: \epsilon_\theta\left(x_t, t\right)
\end{multline}

During training, a neural network $\epsilon_\theta(x_t, t)$ learns to predict the noise $\epsilon_t$ added to a sample $x_t$. Additionally, this network can also be conditioned on text, images, or embeddings~\cite{cfg}, denoted by $\epsilon_\theta(x_t, t, c, \oslash)$, where $c$ is the encoded text condition (using CLIP \cite{clip}) and $\oslash$ is the null condition. In our case, $x_{inv}$ is obtained by providing $c_0$ corresponding to $\mathbf{x}_0$ that is generated using a text-embedding framework such as BLIP~\cite{blip}. \cite{zero-p2p} observe that the inverted noise maps generated by DDIM Inversion $ \epsilon_\theta\left(x_t, t, c, \oslash\right) \in \mathbb{R}^{64 \times 64 \times 4}$ do not follow the statistical properties of uncorrelated, white gaussian noise in most cases, causing poor editability. Thus as in ~\cite{zero-p2p},  we softly enforce gaussianity using a pairwise regularization $\mathcal{L}_{pair}$~\cite{zero-p2p} and a divergence loss $\mathcal{L}_{KL}$~\cite{vae} weighted by $\lambda$. (Details of these losses can be found in Sec. {\color{red}1} of the supplementary material).

Inversion provides us with a good starting point for the editing process, compared to starting from a random latent code (refer Sec. {\color{red}2.1} of the supplementary material for details). However, if we use the standard diffusion process for our edit process, then we will not have control over local regions in the image using simple prompts. To tackle this, we use a MultiDiffusion~\cite{multidiff} process for localized multi-object editing.

\subsection{Diffusion for Multi-Object Editing}

A diffusion model $\Phi$, typically operates as follows: Given a latent code $x_T$ and an encoded prompt $c$, it generates a sequence of latents $\{x_i\}_{i=T-1}^0$ during the backward diffusion process s.t. $x_{t-1} = \Phi(x_t | c)$, gradually denoising $x_T$ over time. To obtain an edited image, we start from $x_T = x_{inv}$ following~\cite{sdedit} and guide it based on a target prompt. This approach applies prompt guidance on the complete image, making the output prone to unintentional edits. Thus, we propose a localized prompting solution, restricting the edits to a masked region.

To concurrently edit $N$ regions corresponding to $N$ masks, one approach is to use $N+1$ different diffusion processes $\{\Phi(x_t^j | c_j)\}_{j=0}^{N}$, where $x_t^j$ and $c_j$ are the latent code and encoded prompts, respectively for mask $j$. However, we adopt a single multidiffusion process~\cite{multidiff} denoted by $\Psi$ for zero-shot conditional editing of regions within all the given $N$ masks. Given masks $\{M_1, \cdots, M_N\}$ and $M_0 = 1 - \bigcup_{i=1}^{N} M_{i}$, with a corresponding set of encoded text prompts $z = (c_1, \cdots, c_N)$, the goal is to come up with a mapping function $ \Psi : \mathcal{X} \times \mathcal{C}^{N+1} \rightarrow \mathcal{X}$, solving the following optimization problem: 
\begin{align}
    \Psi\left(y_t, z\right) = \underset{y_{t-1}}{\operatorname{argmin}} \: \mathcal{L}_{md}(y_{t-1}|y_t, z)
\label{mdp}
\end{align}
A multidiffusion process $\Psi$ starts with $y_T$ and generates a sequence of latents $\{y_i\}_{i=T-1}^0$ given by $y_{t-1} = \Psi(y_t | z)$. The objective in \cref{mdp} is designed to follow the denoising steps of $\Phi$ as closely as possible, enforced using the constraint $\mathcal{L}_{md}$ defined as:
\begin{gather}
    \label{eq:md-constraint}
    \mathcal{L}_{md}(y_{t-1}|y_t, z) =  \sum_{i=0}^N \Big\| M_i \otimes \Big[y_{t-1} - \Phi(x_t^i \: | \: c_i)\Big]  \Big\|^2 
\end{gather}
where $\otimes$ is the Hadamard product. The optimization problem in Eq. \ref{mdp} has a closed-form solution given by:
\begin{equation}
    \Psi\left(y_t, z\right)=\sum_{i=0}^N \frac{M_i}{\sum_{j=0}^N M_j} \otimes \Phi\left(x_t^i \mid c_i\right)   
\end{equation} 

Editing in \lomo is accomplished by running a backward process (termed \textit{edit}), using $\Psi$ with $y_T = x_{inv}$ via a deterministic DDIM reverse process for $\Phi$.
\begin{multline}
x_{t-1}^i=\sqrt{\alpha_{t-1}} \left(\frac{x_t^i - \sqrt{1-\alpha_{t}} \: \epsilon_\theta\left(x_t^i, t, c_i, \oslash\right)}{\sqrt{\alpha_t}}\right)\\+ \sqrt{1-\alpha_{t-1}} \: \epsilon_\theta\left(x_t^i, t, c_i, \oslash\right)
\end{multline}

\par In addition to the \textit{edit} process, we also run a backward process (termed \textit{reconstruction}) using $\Phi$  with $x_T = x_{inv}$ and the source prompt ($c_0$). This provides a reconstruction $x_0^{\prime}$ of the original latent code $x_0$. The deviation of $x_0^{\prime}$ from $x_0$ is rectified by storing noise latents during the inversion process as in \cite{direct-inversion}. During reconstruction, we save the latents $x_t^\prime$ and cross-attention maps $\bar{A}_t$ (\cref{sec:attn}) for all timesteps $t$. These stored latents and attention maps are used to define losses (\cref{sec:losses}) that guide the \textit{edit}.


\subsubsection{Bootstrapping}

As in~\cite{multidiff}, we use a bootstrap parameter ($T_b$), allowing $\Psi(y_t | c_i)$ to focus on region $M_i$ early on in the process (until timestep $T_b$) and consider the full context in the image later on. This will improve the fidelity of the generated images when there are tight masks. It is introduced via time dependency in $y_t$, given by
\begin{equation}
    y_t = \begin{cases}
                    M_i \cdot y_t + (1 - M_i) \cdot b_t, & \text{if $t < T_b$}\\
                    y_t, & \text{otherwise}
		      \end{cases}
\end{equation}
where $b_t$ serves as a background and is obtained by noising the encoded version of a random image with a constant color to the noise level of timestep $t$, i.e. $b_t = \xi(\mathbf{x})$ where $\mathbf{x} \in \mathcal{X}$ and $\xi$ is the Stable Diffusion encoder.

\begin{figure*}[t]
    \centering
\includegraphics[width=0.98\textwidth]{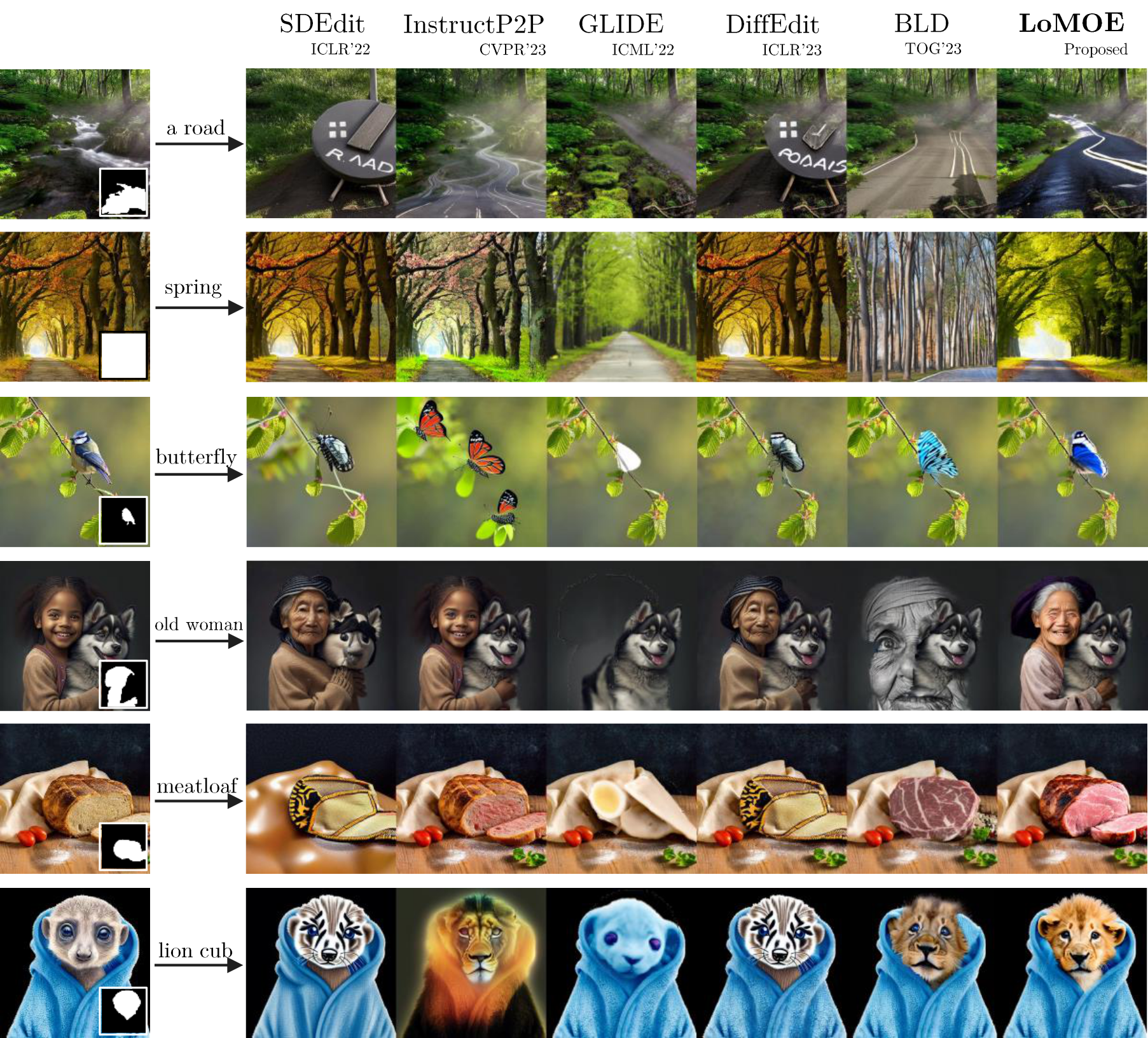}
    \caption{\textbf{Comparison among contemporary methods for Single Object Edits:} We observe that SDEdit~\cite{sdedit} and InstructP2P~\cite{instruct-p2p} tend to modify the whole image. GLIDE~\cite{nichol2022glide} often inpaints and removes the subject of the edit in cases where it fails to generate the edit. DiffEdit~\cite{diffedit} produces the same output as SDEdit while preserving the unmasked regions of the input image. BLD~\cite{bld} doesn't preserve the structure of the input and makes unintented attribute edits to the masked subject. Finally, we observe that our proposed \lomo makes the intented edit, preserves the unmasked region and avoids unintended attribute edits.}
    \label{fig:single-edit-comp}
\end{figure*}

\subsection{Attribute Preservation during Editing}
\label{sec:losses}

While the aforementioned process allows us to solve the multi-object editing problem, it falls short on two accounts (i) maintaining structural consistency with the input image and (ii) reconstructing the background faithfully. We introduce losses $\mathcal{L}_{xa}$ and $\mathcal{L}_{b}$, respectively, to tackle these issues, which are added at each iteration $t$, during the \textit{edit} process.


\subsubsection{Cross-Attention Preservation}
\label{sec:attn}


Diffusion models such as Stable Diffusion~\cite{stable-diff} incorporate cross-attention layers~\cite{aayn} in $\epsilon_\theta$ to effectively condition their generation on text. Throughout the conditional denoising process, the image and text modalities interact with each other for noise prediction. This involves merging the embeddings of visual and textual features through cross-attention layers, yielding spatial attention maps for each textual token.
The attention maps are given by
\begin{equation}
    \bar{A} = \text{Softmax}\left(\frac{Q K^T}{\sqrt{d}}\right)
    \label{eq:attnmap}
\end{equation}
where $Q$ denotes the projection of deep spatial features $\Phi(x_t)$ onto a query matrix $W_Q$, $K$ denotes the projection of the text embedding $c$ onto a key matrix parameterized by $W_K$, and $d$ denotes the latent projection dimension. Note that $\bar{A}_{i,j}$ denotes the weight of the $j^{th}$ text token on the $i^{th}$ pixel. Since the structure and the spatial layout of the generated image depend on $\bar{A}$ ~\cite{prompt2prompt}, during the \textit{edit} process, we update the attention map of the \textit{edit} process $\big(\bar{A}^{e}_{t}\big)$  to follow that of the \textit{reconstruction} process $\big(\bar{A}^{r}_{t}\big)$ via the loss $\mathcal{L}_{xa}$, at each timestep $t$, defined as: $  \mathcal{L}_{xa} = \lVert \bar{A}^{r}_{t} - \bar{A}^{e}_{t} \rVert_{2}$. We also use a temperature parameter~\cite{temp-scaling} $\tau$ in \cref{eq:attnmap} to ensure distributional smoothness (as explained in the supplementary material).


\begin{figure*}
    \centering
\includegraphics[width=0.98\textwidth]{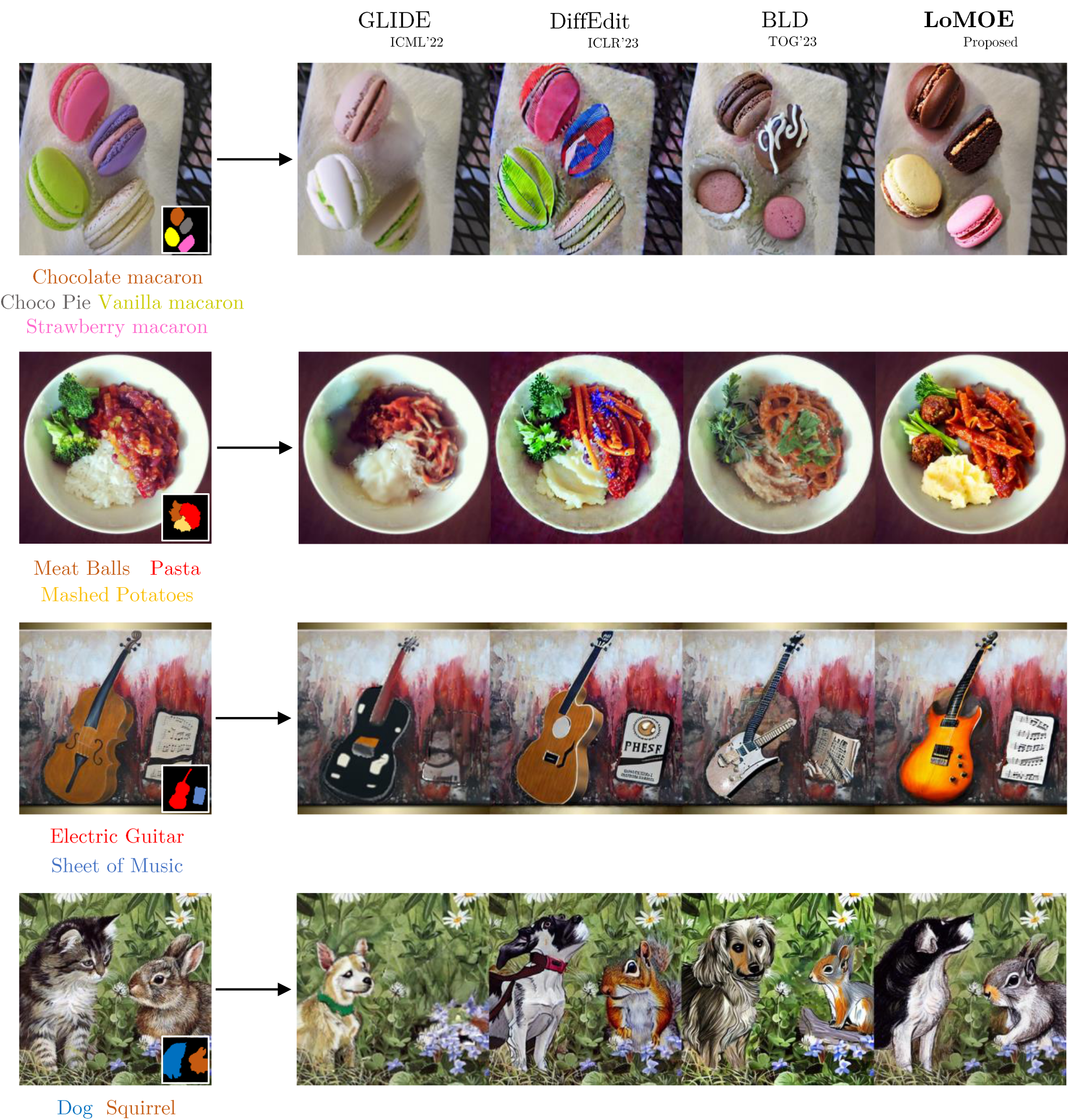}
    \caption{\textbf{Comparison with contemporary methods for Multi-Object Edits:} While the baselines are either unable to make the edit, accumulate artifacts, edit the unmasked region, or make unintended attribute edits,  \lomo is able to faithfully edit in accordance with the prompts.}
    \label{fig:multi-edit-comp}
\end{figure*}

\subsubsection{Background Preservation} To ensure that the \textit{background} of the output stays unchanged, we match the intermediate latents stored during the \textit{reconstruction} process $\left(x_{inv}, \cdots, x_0^{\prime} \right)$ to the latents of the \textit{edit} process $\left(x_{inv}, \cdots, y_0^{*} \right)$. Given the masks $\{M_{i}\}_{i=1}^S{N}$, we obtain the \textit{background} mask as $M_0 = 1 - \bigcup_{i=1}^{N} M_{i}$.  We define a background preservation loss $ \big(\mathcal{L}_{b}\big)$ that makes the background latents of the two processes close to each other, at each timestep $t$ during the edit. Formally, 
   $ \mathcal{L}_{b} = \lVert M_0 \cdot (y_t^{*} - x_t^{\prime}) \rVert_{2}$.  
 Finally, during the \textit{edit} process we use the updated attention maps and latent vector obtained by optimizing $\lambda_{xa} \mathcal{L}_{xa} + \lambda_{b} \mathcal{L}_{b}$, at each time step $t$. Here, $\lambda_{xa}$ and $\lambda_{b}$ are weights of the cross attention and background preservation losses respectively. 

\subsection{Implementation Details}

We employed StableDiffusion v2.0 for our pretrained model $\Phi$. Additionally, we set loss coefficients $\lambda_{b} = 1.75$, $\lambda_{xa} = 1.00$, $\tau = 1.25$ and $T_b = 10$ based on an empirical validation process conducted on a held-out set comprising five images. The majority of our experiments were conducted on a system equipped with a GeForce RTX-3090 with 24 GB of memory. For multi-object edits involving more than five masks, we utilized an A6000 GPU with 48 GB of memory.
\begin{table*}[t]
    \centering
    \begin{tabular}{lcccccccc}
        \toprule
        Method & Mask & \makecell{Source CLIP\\Score ($\uparrow$)} & \makecell{Background\\LPIPS ($\downarrow$)} & \makecell{Background\\PSNR ($\uparrow$)}  & \makecell{Background\\SSIM ($\uparrow$)} & \makecell{Structural\\Distance ($\downarrow$)} & \makecell{Target CLIP\\Score ($\uparrow$)}\\
        \midrule
        Input & - & {25.639} {\footnotesize $\pm$ 0.178} & - & - & - & - & {23.584} {\footnotesize $\pm$ 0.221}\\
        \midrule
        SDEdit \hfill \cite{sdedit} & \xmark & {21.362} {\footnotesize $\pm$ 0.266} & {0.199} {\footnotesize $\pm$ 0.0071} & {24.026} {\footnotesize $\pm$ 0.3269} & {0.788} {\footnotesize $\pm$ 0.0086} & {0.084} {\footnotesize $\pm$ 0.0035} & {23.042} {\footnotesize $\pm$ 0.250}\\
        I-P2P \hfill \cite{instruct-p2p} & \xmark & {22.513} {\footnotesize $\pm$ 0.273} & {0.242} {\footnotesize $\pm$ 0.0123} & {20.406} {\footnotesize $\pm$ 0.4192} & {0.762} {\footnotesize $\pm$ 0.0105} & {0.090} {\footnotesize $\pm$ 0.0042} & {25.038} {\footnotesize $\pm$ 0.216}\\
        GLIDE \hfill \cite{glide} & \cmark & {22.756} {\footnotesize $\pm$ 0.235} & {0.104} {\footnotesize $\pm$ 0.0041} & {\bf 31.798} {\footnotesize $\pm$ 0.2272} & {\bf 0.938} {\footnotesize $\pm$ 0.0031} & {0.094} {\footnotesize $\pm$ 0.0035} & {24.299} {\footnotesize $\pm$ 0.215}\\
        DiffEdit \hfill \cite{diffedit} & \cmark & {\underline{23.269}} {\footnotesize $\pm$ 0.248} & {\underline{0.057}} {\footnotesize $\pm$ 0.0019} & {30.002} {\footnotesize $\pm$ 0.3020} & {0.875} {\footnotesize $\pm$ 0.0063} & {\underline{0.076}} {\footnotesize $\pm$ 0.0036} & {24.094} {\footnotesize $\pm$ 0.234}\\
        BLD \hfill \cite{bld} & \cmark & {22.761} {\footnotesize $\pm$ 0.238} & {0.058} {\footnotesize $\pm$ 0.0021} & {29.389} {\footnotesize $\pm$ 0.2407} & {0.877} {\footnotesize $\pm$ 0.0062} & {0.077} {\footnotesize $\pm$ 0.0034} & {\underline{25.867}} {\footnotesize $\pm$ 0.206}\\
        \midrule
        \lomo & \cmark & {\bf 23.545} {\footnotesize $\pm$ 0.219} & {\bf 0.054} {\footnotesize $\pm$ 0.0022} & {\underline{30.315}} {\footnotesize $\pm$ 0.2558} & {\underline{0.885}} {\footnotesize $\pm$ 0.0060} & {\bf 0.066} {\footnotesize $\pm$ 0.0031} & {\bf 26.074} {\footnotesize $\pm$ 0.201} \\
        \bottomrule
    \end{tabular}
    \caption{\textbf{Comparison with different baselines for Single-Object Edits:} We use a large array of \textit{classical} and \textit{neural} metrics that provides valuable statistical insights regarding the edit properties of considered methods. The best performing method is indicated using \textbf{bold} and the second best is \underline{underlined}. In particular, \lomo outperforms on all \textit{neural} metrics indicating realistic image generation. \lomo also performs faithful edits indicated by high \textit{classical} metrics.}
    \label{tab:single-edit-experiments} 
    \vspace{-0.5em}
\end{table*}

\section{Experimental Setting}

We consider two sets of experiments: (a) single-object edits and (b) multi-object edits. For the multi-object editing experiments, while \lomo can be employed as it is, we resort to iterative editing for other methods. We report both qualitative and quantitative outcomes of our experiments.

\subsection{Datasets}

For single-object edits, we utilized a modified subset of the PIE-Bench~\cite{direct-inversion} dataset, supplemented with images from AFHQ~\cite{animal-faces}, COCO~\cite{coco}, and Imagen~\cite{imagen}. For multi-object edits, we introduce a new dataset named \proposedDataset, comprising 64 images featuring 2 to 7 masks, each paired with corresponding text prompts. The details of the curated dataset can be found in the supplementary material. The \proposedDataset dataset will be made public in due time.

\subsection{Baseline Methods}

We benchmark \lomo against several methods, including SDEdit~\cite{sdedit}, Instruct-Pix2Pix (I-P2P)~\cite{instruct-p2p}, GLIDE~\cite{glide}, DiffEdit~\cite{diffedit} and Blended Latent Diffusion (BLD)~\cite{bld}. Official implementations were used for all methods, except for SDEdit and DiffEdit.
GLIDE, DiffEdit, BLD, and \lomo leverage masks, whereas the other methods operate on the whole image. Additionally, there are differences among the methods in terms of the types of text prompts they require. SDEdit and DiffEdit necessitate both source and target text prompts, and I-P2P takes edit instructions as prompts, prompting us to extend PIE-Bench to accommodate these methods. Similar to \lomo, both GLIDE and BLD only use edit prompts corresponding to the masks. Finally, given the considerably noisy masks generated by DiffEdit, we opted to provide it with ground truth masks.

\subsection{Metrics}
We quantitatively analyze the edited images on a set of \textit{neural} metrics, namely Clip Score (CS)~\cite{clipscore} with both source and target prompts, Background (BG)-LPIPS~\cite{lpips}, and Structural Distance~\cite{strucdist}. Additionally, we employed \textit{classical} metrics, including BG-PSNR and BG-SSIM. The \textit{neural} metrics evaluate the perceptual similarity of the image, emphasizing realism. On the other hand, \textit{classical} metrics focus on pixel-level similarity and don't comment on the realism or quality of the edit. In contrast to previous methods, we introduce comparisons over a new target CS metric. We also provide target prompts for all images in both datasets, enabling a more effective measure of the edit quality.  To ensure robustness in our assessments, we averaged all the metrics over 5 seeds and reported the average standard error for all methods. Additionally, we conduct a subjective evaluation experiment to assess the quality of edits, described in the supplementary material.

\section{Results and Discussion}
\begin{table*}[t]
    \centering
    \begin{tabular}{lcccccccc}
        \toprule
        Method & \makecell{Single\\Pass} & \makecell{Source CLIP\\Score ($\uparrow$)} & \makecell{Background\\LPIPS ($\downarrow$)} & \makecell{Background\\PSNR ($\uparrow$)}  & \makecell{Background\\SSIM ($\uparrow$)} & \makecell{Structural\\Distance ($\downarrow$)} & \makecell{Target CLIP\\Score ($\uparrow$)}\\
        \midrule
        Input & - & {26.956} {\footnotesize $\pm$ 0.141} & - & - & - & - & {22.489} {\footnotesize $\pm$ 0.236}\\
        \midrule
        GLIDE \hfill \cite{glide} & \xmark & {\bf 27.038} {\footnotesize $\pm$ 0.308} & {0.192} {\footnotesize $\pm$ 0.0151} & {\bf 30.196} {\footnotesize $\pm$ 0.4748} & {\bf 0.894} {\footnotesize $\pm$ 0.0104} & {0.085} {\footnotesize $\pm$ 0.0065} & {22.754} {\footnotesize $\pm$ 0.526}\\
        DiffEdit \hfill \cite{diffedit} & \xmark & {\underline{26.417}} {\footnotesize $\pm$ 0.306} & {0.188} {\footnotesize $\pm$ 0.0119} & {24.559} {\footnotesize $\pm$ 0.4528} & {0.756} {\footnotesize $\pm$ 0.0168} & {\underline{0.071}} {\footnotesize $\pm$ 0.0063} & {23.898} {\footnotesize $\pm$ 0.445}\\
        BLD \hfill \cite{bld} & \xmark & {26.330} {\footnotesize $\pm$ 0.268} & {\underline{0.126}} {\footnotesize $\pm$ 0.0086} & {26.632} {\footnotesize $\pm$ 0.4627} & {0.800} {\footnotesize $\pm$ 0.0150} & {0.074} {\footnotesize $\pm$ 0.0062} & {\underline{25.394}} {\footnotesize $\pm$ 0.450}\\
        \midrule
        \lomo & \cmark & {25.959} {\footnotesize $\pm$ 0.111} & {\bf 0.107} {\footnotesize $\pm$ 0.0040} & {\underline{27.222}} {\footnotesize $\pm$ 0.2053} & {\underline{0.826}} {\footnotesize $\pm$ 0.0073} & {\bf 0.066} {\footnotesize $\pm$ 0.0027} & {\bf 26.154} {\footnotesize $\pm$ 0.187} \\
        \bottomrule
    \end{tabular}
    \caption{\textbf{Comparison with different baselines for Multi-Object Edits:} We use a large array of \textit{classical} and \textit{neural} metrics that provides valuable statistical insights regarding the edit properties of considered methods. The best performing method is indicated using \textbf{bold} and the second best is \underline{underlined}. We observe that only \lomo has a higher target CS compared to source CS.}
    \label{tab:multi-edit-experiments}
    \vspace{-0.5em}
\end{table*}
\begin{table}[t]
    \adjustbox{max width=\linewidth}{%
    \centering
    \begin{tabular}{ccccc}
        \toprule
        $\mathcal{L}_{xa}$ & $\mathcal{L}_{b}$ & \makecell{Source CLIP\\Score ($\uparrow$)} & \makecell{Structural\\Distance ($\downarrow$)} & \makecell{Target CLIP\\Score ($\uparrow$)}\\
        \midrule
        \xmark & \xmark & 23.0906 & 0.0763 & 26.2555\\
        \xmark & \cmark & 23.3925 & 0.0728 & \textbf{26.2662}\\
        \cmark & \xmark & \textbf{23.6611} & 0.0699 & 26.1338\\
        \cmark & \cmark & 23.5445 & \textbf{0.0661} & 26.0740\\
        \midrule
        \midrule
        $\mathcal{L}_{xa}$ & $\mathcal{L}_{b}$ & \makecell{Background\\LPIPS ($\downarrow$)} & \makecell{Background\\PSNR ($\uparrow$)}  & \makecell{Background\\SSIM ($\uparrow$)}\\
        \midrule
        \xmark & \xmark & 0.1088 & 26.4474 & 0.8537\\
        \xmark & \cmark & 0.0554 & 30.1475 & 0.8818\\
        \cmark & \xmark & 0.0749 & 26.9587 & 0.8698\\
        \cmark & \cmark & \textbf{0.0546} & \textbf{30.3154} & \textbf{0.8847}\\
        \bottomrule
    \end{tabular}}
    \caption{\textbf{Ablation Study:} We observe that both our losses complement each other and result in improved metrics.}
    \label{tab:ablation-experiments}
    \vspace{-0.5em}
\end{table}

\subsection{Single Object Edits}

In  this case, \lomo offers better \textit{neural} metrics compared to all the baselines (\cref{tab:single-edit-experiments}). This attests to \lomo's adeptness in executing edits while maintaining fidelity to the source image and prompt. However, in terms of \textit{classical} metrics, GLIDE outperforms \lomo, revealing a trade-off between realism and faithfulness, akin to observations in Meng \etal~\cite{sdedit}. GLIDE excels in \textit{classical} metrics due to its inherent inpainting model design, but it lags in \textit{neural} metrics, resulting in less realistic images. BLD and I-P2P exhibit good target CS metrics but lag behind in other aspects. Particularly, I-P2P demonstrates subpar BG metrics, attributed to its operation on the entire image without the use of a mask. Notably, instances where the target CS closely aligns with that of the Input (no edit) suggest the absence of applied edits.  Figure \ref{fig:single-edit-comp} depicts a few examples of all the compared methods with \lomo producing visually faithful edits. 

\subsection{Multi-Object Edits}

Similar to our observations in single-object editing, \lomo exhibits superior performance across all \textit{neural} metrics in multi-object editing, except for source CS. This deviation is anticipated, given the substantial image transformations in multi-object editing. Ideally, such transformations lead to images that are markedly different from the source prompt and more aligned with the target prompt. Therefore, elevated BG-LPIPS and Structural Distance better indicate perceptual quality, while a high target CS signifies successful editing. Conversely, all other methods display a considerably lower target CS compared to source CS, indicating unsuccessful edits. We also note a trade-off between preservation and editing in multi-object scenarios. Intuitively, as the number of edited objects increases, the source CS tends to decrease, while the target CS tends to increase. Furthermore, given our single-pass approach, we achieve significant savings in edit time compared to methods that perform multi-edits iteratively. Additional details can be found in the supplementary material. Figure \ref{fig:multi-edit-comp} shows qualitative results on all the compared methods on a few sample images which demonstrate \lomo's impressive performance in preserving the intricate details during edits. 

\subsection{Ablation Studies}

To assess the significance of each loss component in \lomo, we conducted a comprehensive ablation study, maintaining a fixed seed, $\tau$ and $T_b$. Detailed ablation results for varying values of $\tau$ and $T_b$, along with limitations, can be found in the supplementary material. The findings presented in Table \ref{tab:ablation-experiments} reveal that incorporating $\mathcal{L}_{xa}$ enhances \textit{neural} metrics, contributing to the realism of the edited image. Meanwhile, the inclusion of $\mathcal{L}_{b}$ improves our \textit{classical} metrics, enhancing the faithfulness of the edited image. Notably, these two aspects - realism and faithfulness are orthogonal qualities in image generation and editing. The combination of both losses in \lomo yields improved performance, achieving a balanced enhancement in both the realism and faithfulness of the edit.
\section{Conclusion}

We present \lomo, a framework designed to address the challenging task of localized multi-object editing using diffusion models. Our approach enables (mask and prompt)-driven multi-object editing without the need for prior training, allowing diverse operations on complex scenes in a single pass, thereby having improved inference speed compared to iterative single-object editing methods. Our framework achieves high-quality reconstructions with minimal artifacts through cross-attention and background preservation losses. Further, we curate \proposedDataset, a benchmark dataset that provides a valuable platform for evaluating multi-object image editing frameworks. Experimental evaluations demonstrate \lomo's superior performance in both image editing quality and faithfulness compared to current benchmarks. We believe that \lomo would serve as an effective tool for artists and designers.

{
    \small
    \bibliographystyle{ieeenat_fullname}
    \bibliography{main}

\begin{thebibliography}{48}
\providecommand{\natexlab}[1]{#1}
\providecommand{\url}[1]{\texttt{#1}}
\expandafter\ifx\csname urlstyle\endcsname\relax
  \providecommand{\doi}[1]{doi: #1}\else
  \providecommand{\doi}{doi: \begingroup \urlstyle{rm}\Url}\fi

\bibitem[Andonian et~al.(2021)Andonian, Osmany, Cui, Park, Jahanian, Torralba, and Bau]{paint-by-word}
Alex Andonian, Sabrina Osmany, Audrey Cui, YeonHwan Park, Ali Jahanian, Antonio Torralba, and David Bau.
\newblock Paint by word, 2021.

\bibitem[Avrahami et~al.(2023)Avrahami, Fried, and Lischinski]{bld}
Omri Avrahami, Ohad Fried, and Dani Lischinski.
\newblock Blended latent diffusion.
\newblock \emph{ACM Trans. Graph.}, 42\penalty0 (4), 2023.

\bibitem[Bar-Tal et~al.(2023)Bar-Tal, Yariv, Lipman, and Dekel]{multidiff}
Omer Bar-Tal, Lior Yariv, Yaron Lipman, and Tali Dekel.
\newblock Multidiffusion: Fusing diffusion paths for controlled image generation.
\newblock In \emph{ICML}. PMLR, 2023.

\bibitem[Bau et~al.(2019)Bau, Zhu, Wulff, Peebles, Strobelt, Zhou, and Torralba]{gan-cannot}
David Bau, Jun-Yan Zhu, Jonas Wulff, William Peebles, Hendrik Strobelt, Bolei Zhou, and Antonio Torralba.
\newblock Seeing what a gan cannot generate.
\newblock In \emph{Proceedings of the IEEE/CVF International Conference on Computer Vision}, pages 4502--4511, 2019.

\bibitem[Brooks et~al.(2023)Brooks, Holynski, and Efros]{instruct-p2p}
Tim Brooks, Aleksander Holynski, and Alexei~A Efros.
\newblock Instructpix2pix: Learning to follow image editing instructions.
\newblock In \emph{Proceedings of the IEEE/CVF Conference on Computer Vision and Pattern Recognition}, pages 18392--18402, 2023.

\bibitem[Choi et~al.(2020)Choi, Uh, Yoo, and Ha]{animal-faces}
Yunjey Choi, Youngjung Uh, Jaejun Yoo, and Jung-Woo Ha.
\newblock Stargan v2: Diverse image synthesis for multiple domains.
\newblock In \emph{Proceedings of the IEEE Conference on Computer Vision and Pattern Recognition}, 2020.

\bibitem[Christmas et~al.(1995)Christmas, Kittler, and Petrou]{strucdist}
W.J. Christmas, J. Kittler, and M. Petrou.
\newblock Structural matching in computer vision using probabilistic relaxation.
\newblock \emph{IEEE Transactions on Pattern Analysis and Machine Intelligence}, 17\penalty0 (8):\penalty0 749--764, 1995.

\bibitem[Couairon et~al.(2023)Couairon, Verbeek, Schwenk, and Cord]{diffedit}
Guillaume Couairon, Jakob Verbeek, Holger Schwenk, and Matthieu Cord.
\newblock Diffedit: Diffusion-based semantic image editing with mask guidance.
\newblock In \emph{The Eleventh International Conference on Learning Representations}, 2023.

\bibitem[Crowson et~al.(2022)Crowson, Biderman, Kornis, Stander, Hallahan, Castricato, and Raff]{vqgan-clip}
Katherine Crowson, Stella Biderman, Daniel Kornis, Dashiell Stander, Eric Hallahan, Louis Castricato, and Edward Raff.
\newblock Vqgan-clip: Open domain image generation and editing with natural language guidance.
\newblock In \emph{European Conference on Computer Vision}, pages 88--105. Springer, 2022.

\bibitem[Dhariwal and Nichol(2021)]{dhariwal2021diffusion}
Prafulla Dhariwal and Alexander Nichol.
\newblock Diffusion models beat gans on image synthesis.
\newblock \emph{Advances in neural information processing systems}, 34:\penalty0 8780--8794, 2021.

\bibitem[Gafni et~al.(2022)Gafni, Polyak, Ashual, Sheynin, Parikh, and Taigman]{make-a-scene}
Oran Gafni, Adam Polyak, Oron Ashual, Shelly Sheynin, Devi Parikh, and Yaniv Taigman.
\newblock Make-a-scene: Scene-based text-to-image generation with human priors.
\newblock In \emph{European Conference on Computer Vision}, pages 89--106. Springer, 2022.

\bibitem[Gal et~al.(2022)Gal, Patashnik, Maron, Bermano, Chechik, and Cohen-Or]{gal2022stylegan}
Rinon Gal, Or Patashnik, Haggai Maron, Amit~H Bermano, Gal Chechik, and Daniel Cohen-Or.
\newblock Stylegan-nada: Clip-guided domain adaptation of image generators.
\newblock \emph{ACM Transactions on Graphics (TOG)}, 41\penalty0 (4):\penalty0 1--13, 2022.

\bibitem[Goodfellow et~al.(2014)Goodfellow, Pouget-Abadie, Mirza, Xu, Warde-Farley, Ozair, Courville, and Bengio]{goodfellow2014generative}
Ian Goodfellow, Jean Pouget-Abadie, Mehdi Mirza, Bing Xu, David Warde-Farley, Sherjil Ozair, Aaron Courville, and Yoshua Bengio.
\newblock Generative adversarial nets.
\newblock \emph{Advances in neural information processing systems}, 27, 2014.

\bibitem[Guo et~al.(2019)Guo, Pleiss, Sun, and Weinberger]{temp-scaling}
Chuan Guo, Geoff Pleiss, Yu Sun, and Kilian~Q Weinberger.
\newblock On calibration of modern neural networks.
\newblock 2019.

\bibitem[He et~al.(2023)He, Salakhutdinov, and Kolter]{localized-compose}
Yutong He, Ruslan Salakhutdinov, and J~Zico Kolter.
\newblock Localized text-to-image generation for free via cross attention control.
\newblock \emph{arXiv preprint arXiv:2306.14636}, 2023.

\bibitem[Hein et~al.(2018)Hein, Andriushchenko, and Bitterwolf]{temp-scaling-calib}
Matthias Hein, Maksym Andriushchenko, and Julian Bitterwolf.
\newblock Why relu networks yield high-confidence predictions far away from the training data and how to mitigate the problem.
\newblock \emph{2019 IEEE/CVF Conference on Computer Vision and Pattern Recognition (CVPR)}, pages 41--50, 2018.

\bibitem[Hertz et~al.(2022)Hertz, Mokady, Tenenbaum, Aberman, Pritch, and Cohen-Or]{prompt2prompt}
Amir Hertz, Ron Mokady, Jay Tenenbaum, Kfir Aberman, Yael Pritch, and Daniel Cohen-Or.
\newblock Prompt-to-prompt image editing with cross attention control.
\newblock 2022.

\bibitem[Hessel et~al.(2021)Hessel, Holtzman, Forbes, Bras, and Choi]{clipscore}
Jack Hessel, Ari Holtzman, Maxwell Forbes, Ronan~Le Bras, and Yejin Choi.
\newblock Clipscore: A reference-free evaluation metric for image captioning.
\newblock \emph{arXiv preprint arXiv:2104.08718}, 2021.

\bibitem[Ho and Salimans(2021)]{cfg}
Jonathan Ho and Tim Salimans.
\newblock Classifier-free diffusion guidance.
\newblock In \emph{NeurIPS 2021 Workshop on Deep Generative Models and Downstream Applications}, 2021.

\bibitem[Ho et~al.(2020)Ho, Jain, and Abbeel]{ddpm}
Jonathan Ho, Ajay Jain, and Pieter Abbeel.
\newblock Denoising diffusion probabilistic models.
\newblock \emph{Advances in neural information processing systems}, 33:\penalty0 6840--6851, 2020.

\bibitem[Ju et~al.(2023)Ju, Zeng, Bian, Liu, and Xu]{direct-inversion}
Xuan Ju, Ailing Zeng, Yuxuan Bian, Shaoteng Liu, and Qiang Xu.
\newblock Direct inversion: Boosting diffusion-based editing with 3 lines of code.
\newblock \emph{arXiv preprint arXiv:2304.04269}, 2023.

\bibitem[Karras et~al.(2020)Karras, Laine, Aittala, Hellsten, Lehtinen, and Aila]{style-gan}
T. Karras, S. Laine, M. Aittala, J. Hellsten, J. Lehtinen, and T. Aila.
\newblock Analyzing and improving the image quality of stylegan.
\newblock In \emph{2020 IEEE/CVF Conference on Computer Vision and Pattern Recognition (CVPR)}, pages 8107--8116, Los Alamitos, CA, USA, 2020. IEEE Computer Society.

\bibitem[Kawar et~al.(2023)Kawar, Zada, Lang, Tov, Chang, Dekel, Mosseri, and Irani]{imagic}
Bahjat Kawar, Shiran Zada, Oran Lang, Omer Tov, Huiwen Chang, Tali Dekel, Inbar Mosseri, and Michal Irani.
\newblock Imagic: Text-based real image editing with diffusion models.
\newblock In \emph{Proceedings of the IEEE/CVF Conference on Computer Vision and Pattern Recognition}, pages 6007--6017, 2023.

\bibitem[Kim et~al.(2023)Kim, Lee, Kim, Ha, and Zhu]{dense-t2i-compose}
Yunji Kim, Jiyoung Lee, Jin-Hwa Kim, Jung-Woo Ha, and Jun-Yan Zhu.
\newblock Dense text-to-image generation with attention modulation.
\newblock In \emph{Proceedings of the IEEE/CVF International Conference on Computer Vision}, pages 7701--7711, 2023.

\bibitem[Kingma and Welling(2014)]{vae}
Diederik~P. Kingma and Max Welling.
\newblock Auto-encoding variational bayes.
\newblock In \emph{International Conference on Learning Representations, {ICLR} 2014}, 2014.

\bibitem[Kirillov et~al.(2023)Kirillov, Mintun, Ravi, Mao, Rolland, Gustafson, Xiao, Whitehead, Berg, Lo, Doll{\'a}r, and Girshick]{sam}
Alexander Kirillov, Eric Mintun, Nikhila Ravi, Hanzi Mao, Chloe Rolland, Laura Gustafson, Tete Xiao, Spencer Whitehead, Alexander~C. Berg, Wan-Yen Lo, Piotr Doll{\'a}r, and Ross Girshick.
\newblock Segment anything.
\newblock \emph{arXiv:2304.02643}, 2023.

\bibitem[Li et~al.(2022)Li, Li, Xiong, and Hoi]{blip}
Junnan Li, Dongxu Li, Caiming Xiong, and Steven Hoi.
\newblock Blip: Bootstrapping language-image pre-training for unified vision-language understanding and generation.
\newblock In \emph{ICML}, 2022.

\bibitem[Li et~al.(2023)Li, Liu, Wu, Mu, Yang, Gao, Li, and Lee]{gligen}
Yuheng Li, Haotian Liu, Qingyang Wu, Fangzhou Mu, Jianwei Yang, Jianfeng Gao, Chunyuan Li, and Yong~Jae Lee.
\newblock Gligen: Open-set grounded text-to-image generation.
\newblock \emph{CVPR}, 2023.

\bibitem[Lin et~al.(2014)Lin, Maire, Belongie, Hays, Perona, Ramanan, Doll{\'a}r, and Zitnick]{coco}
Tsung-Yi Lin, Michael Maire, Serge Belongie, James Hays, Pietro Perona, Deva Ramanan, Piotr Doll{\'a}r, and C~Lawrence Zitnick.
\newblock Microsoft coco: Common objects in context.
\newblock In \emph{Computer Vision--ECCV 2014: 13th European Conference, Zurich, Switzerland, September 6-12, 2014, Proceedings, Part V 13}, pages 740--755. Springer, 2014.

\bibitem[Lu et~al.(2022)Lu, Zhou, Bao, Chen, Li, and Zhu]{dpmsolver}
Cheng Lu, Yuhao Zhou, Fan Bao, Jianfei Chen, Chongxuan Li, and Jun Zhu.
\newblock Dpm-solver: A fast ode solver for diffusion probabilistic model sampling in around 10 steps.
\newblock \emph{arXiv preprint arXiv:2206.00927}, 2022.

\bibitem[Mansimov et~al.(2015)Mansimov, Parisotto, Ba, and Salakhutdinov]{rnn}
Elman Mansimov, Emilio Parisotto, Jimmy~Lei Ba, and Ruslan Salakhutdinov.
\newblock Generating images from captions with attention.
\newblock \emph{arXiv preprint arXiv:1511.02793}, 2015.

\bibitem[Meng et~al.(2021)Meng, He, Song, Song, Wu, Zhu, and Ermon]{sdedit}
Chenlin Meng, Yutong He, Yang Song, Jiaming Song, Jiajun Wu, Jun-Yan Zhu, and Stefano Ermon.
\newblock Sdedit: Guided image synthesis and editing with stochastic differential equations.
\newblock \emph{arXiv preprint arXiv:2108.01073}, 2021.

\bibitem[Mokady et~al.(2023)Mokady, Hertz, Aberman, Pritch, and Cohen-Or]{null-text}
Ron Mokady, Amir Hertz, Kfir Aberman, Yael Pritch, and Daniel Cohen-Or.
\newblock Null-text inversion for editing real images using guided diffusion models.
\newblock In \emph{Proceedings of the IEEE/CVF Conference on Computer Vision and Pattern Recognition}, pages 6038--6047, 2023.

\bibitem[Nichol et~al.(2022{\natexlab{a}})Nichol, Dhariwal, Ramesh, Shyam, Mishkin, Mcgrew, Sutskever, and Chen]{glide}
Alexander~Quinn Nichol, Prafulla Dhariwal, Aditya Ramesh, Pranav Shyam, Pamela Mishkin, Bob Mcgrew, Ilya Sutskever, and Mark Chen.
\newblock Glide: Towards photorealistic image generation and editing with text-guided diffusion models.
\newblock In \emph{International Conference on Machine Learning}, pages 16784--16804. PMLR, 2022{\natexlab{a}}.

\bibitem[Nichol et~al.(2022{\natexlab{b}})Nichol, Dhariwal, Ramesh, Shyam, Mishkin, Mcgrew, Sutskever, and Chen]{nichol2022glide}
Alexander~Quinn Nichol, Prafulla Dhariwal, Aditya Ramesh, Pranav Shyam, Pamela Mishkin, Bob Mcgrew, Ilya Sutskever, and Mark Chen.
\newblock Glide: Towards photorealistic image generation and editing with text-guided diffusion models.
\newblock In \emph{International Conference on Machine Learning}, pages 16784--16804. PMLR, 2022{\natexlab{b}}.

\bibitem[Paiss et~al.(2022)Paiss, Chefer, and Wolf]{clip-gen2}
Roni Paiss, Hila Chefer, and Lior Wolf.
\newblock No token left behind: Explainability-aided image classification and generation.
\newblock In \emph{European Conference on Computer Vision}, pages 334--350. Springer, 2022.

\bibitem[Parmar et~al.(2023)Parmar, Kumar~Singh, Zhang, Li, Lu, and Zhu]{zero-p2p}
Gaurav Parmar, Krishna Kumar~Singh, Richard Zhang, Yijun Li, Jingwan Lu, and Jun-Yan Zhu.
\newblock Zero-shot image-to-image translation.
\newblock In \emph{ACM SIGGRAPH 2023 Conference Proceedings}, pages 1--11, 2023.

\bibitem[Pl\"{o}tz and Roth(2018)]{temp-scaling-restoration}
Tobias Pl\"{o}tz and Stefan Roth.
\newblock Neural nearest neighbors networks.
\newblock In \emph{Proceedings of the 32nd International Conference on Neural Information Processing Systems}, page 1095–1106, 2018.

\bibitem[Radford et~al.(2021)Radford, Kim, Hallacy, Ramesh, Goh, Agarwal, Sastry, Askell, Mishkin, Clark, et~al.]{clip}
Alec Radford, Jong~Wook Kim, Chris Hallacy, Aditya Ramesh, Gabriel Goh, Sandhini Agarwal, Girish Sastry, Amanda Askell, Pamela Mishkin, Jack Clark, et~al.
\newblock Learning transferable visual models from natural language supervision.
\newblock In \emph{International conference on machine learning}, pages 8748--8763. PMLR, 2021.

\bibitem[Ramesh et~al.(2021)Ramesh, Pavlov, Goh, Gray, Voss, Radford, Chen, and Sutskever]{dall-e}
Aditya Ramesh, Mikhail Pavlov, Gabriel Goh, Scott Gray, Chelsea Voss, Alec Radford, Mark Chen, and Ilya Sutskever.
\newblock Zero-shot text-to-image generation.
\newblock In \emph{International Conference on Machine Learning}, pages 8821--8831. PMLR, 2021.

\bibitem[Rombach et~al.(2022)Rombach, Blattmann, Lorenz, Esser, and Ommer]{stable-diff}
Robin Rombach, Andreas Blattmann, Dominik Lorenz, Patrick Esser, and Bj{\"o}rn Ommer.
\newblock High-resolution image synthesis with latent diffusion models.
\newblock In \emph{Proceedings of the IEEE/CVF conference on computer vision and pattern recognition}, pages 10684--10695, 2022.

\bibitem[Song et~al.(2021)Song, Meng, and Ermon]{ddim}
Jiaming Song, Chenlin Meng, and Stefano Ermon.
\newblock Denoising diffusion implicit models.
\newblock In \emph{International Conference on Learning Representations}, 2021.

\bibitem[Vaswani et~al.(2017)Vaswani, Shazeer, Parmar, Uszkoreit, Jones, Gomez, Kaiser, and Polosukhin]{aayn}
Ashish Vaswani, Noam Shazeer, Niki Parmar, Jakob Uszkoreit, Llion Jones, Aidan~N Gomez, \L~ukasz Kaiser, and Illia Polosukhin.
\newblock Attention is all you need.
\newblock In \emph{Advances in Neural Information Processing Systems}. Curran Associates, Inc., 2017.

\bibitem[Wang et~al.(2023)Wang, Saharia, Montgomery, Pont-Tuset, Noy, Pellegrini, Onoe, Laszlo, Fleet, Soricut, et~al.]{imagen}
Su Wang, Chitwan Saharia, Ceslee Montgomery, Jordi Pont-Tuset, Shai Noy, Stefano Pellegrini, Yasumasa Onoe, Sarah Laszlo, David~J Fleet, Radu Soricut, et~al.
\newblock Imagen editor and editbench: Advancing and evaluating text-guided image inpainting.
\newblock In \emph{Proceedings of the IEEE/CVF Conference on Computer Vision and Pattern Recognition}, pages 18359--18369, 2023.

\bibitem[Wang et~al.(2022)Wang, Liu, He, Wu, and Yi]{wang2022clip-gen}
Zihao Wang, Wei Liu, Qian He, Xinglong Wu, and Zili Yi.
\newblock Clip-gen: Language-free training of a text-to-image generator with clip.
\newblock \emph{arXiv preprint arXiv:2203.00386}, 2022.

\bibitem[Zhang et~al.(2023)Zhang, Rao, and Agrawala]{controlnet}
Lvmin Zhang, Anyi Rao, and Maneesh Agrawala.
\newblock Adding conditional control to text-to-image diffusion models.
\newblock In \emph{Proceedings of the IEEE/CVF International Conference on Computer Vision}, pages 3836--3847, 2023.

\bibitem[Zhang et~al.(2018)Zhang, Isola, Efros, Shechtman, and Wang]{lpips}
Richard Zhang, Phillip Isola, Alexei~A. Efros, Eli Shechtman, and Oliver Wang.
\newblock The unreasonable effectiveness of deep features as a perceptual metric.
\newblock In \emph{Proceedings of the IEEE Conference on Computer Vision and Pattern Recognition (CVPR)}, 2018.

\bibitem[Zhou et~al.(2023)Zhou, Zeng, and Gong]{temp-scaling-inpainting}
Xiang Zhou, Yuan Zeng, and Yi Gong.
\newblock Learning to scale temperature in masked self-attention for image inpainting.
\newblock \emph{ArXiv}, abs/2302.06130, 2023.

\end{thebibliography}
}

\addtocontents{toc}{\protect\setcounter{tocdepth}{3}}
\clearpage
\maketitlesupplementary

\begingroup
\let\clearpage\relax
\hypersetup{linkcolor=blue}
\tableofcontents
\endgroup

\setcounter{page}{1}
\setcounter{section}{0}

\section{Introduction}

To keep the overall manuscript self-contained, we include additional details in the supplementary material. The source code for \lomo along with the \proposedDataset dataset will be released in due time.

\section{Method Details}

Specific aspects of the framework, including regularized inversion and temperature scaling, are described below.

\begin{figure*}[t]
    \centering
\includegraphics[width=\textwidth]{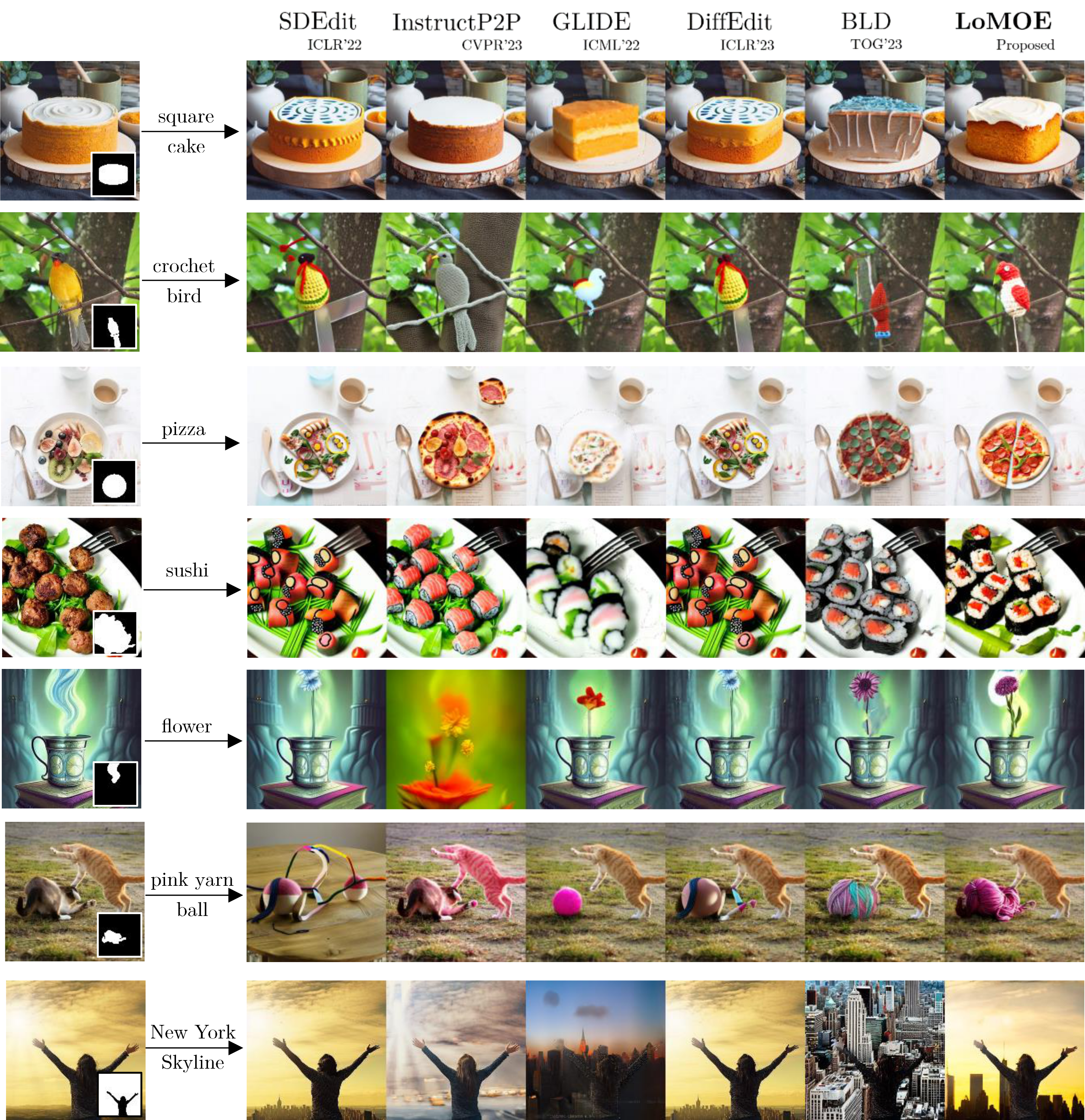}
    \caption{\textbf{Additional Comparison among Contemporary Methods for Single Object Edits:} We present a qualitative comparison of \lomo against other baseline methods on additional single-object edits. The observations stand similar to Fig. {\color{red}3} in the main paper, where our proposed method \lomo makes the intented edit, preserves the unmasked region and avoids unintended attribute edits.}
    \label{fig:single-edit-supp}
\end{figure*}

\subsection{Regularised Inversion}

To softly enforce gaussianity on the inverted noise maps generated during the DDIM Inversion, we use a pairwise regularization $\mathcal{L}_{pair}$~\cite{zero-p2p} and a divergence loss $\mathcal{L}_{KL}$~\cite{vae} weighted by $\lambda$ (refer Sec. {\color{red}3.1} of the main paper). These losses ensure that there is \textbf{(1)} no correlation between any pair of random locations and \textbf{(2)} zero mean, unit variance at each spatial location, respectively. Mathematically, the pairwise regularization loss is given by:

\begin{equation}
\mathcal{L}_{\text{pair}}=\sum_p \frac{1}{S_p^2} \sum_{\delta=1}^{S_p-1} \sum_{x, y, c} \eta_{x, y, c}^p\left(\eta_{x-\delta, y, c}^p+\eta_{x, y-\delta, c}^p\right)
\end{equation}
where $\{\eta^{0}, \eta^{1}, \cdots, \eta^{p}\}$ denote the noise maps with size $S_p$ at the $p$\textsuperscript{th} pyramid level, $\delta$ denotes the offset which helps propagate long-range information \cite{style-gan,zero-p2p}, and $\{x,y,c\}$ denotes a spatial location. Here, we set $p=4$ and $\eta^{0} = \epsilon_\theta \in \mathbb{R}^{64 \times 64 \times 4}$, where the subsequent noise maps are obtained via max-pooling.

\noindent
The divergence loss is given by:
\begin{equation}
    \mathcal{L}_{KL} = \sigma_{\epsilon_\theta}^2 + \mu_{\epsilon_\theta}^2 - 1 - \log(\sigma_{\epsilon_\theta}^2 + \varepsilon)
\end{equation}
where $\mu_{\epsilon_\theta}$ and $\sigma^{2}_{\epsilon_\theta}$ denotes the mean and variance of ${\epsilon_\theta}$ and $\varepsilon$ is a stabilization constant.

\subsection{Temperature Scaling}

Given a vector $z = (z_1, \cdots, z_n) \in \mathbb{R}^{n}$, it can be transformed into a probability vector via
\begin{equation}
   \text{Softmax}(z|\tau)_{i} = \frac{e^{z_i/\tau}}{\sum_{j=1}^{n} e^{z_j/\tau}}
\end{equation}
where $\tau$ is a temperature parameter~\cite{temp-scaling} which varies the smoothness of the output distribution. In general, lower values of $\tau$ result in a sharp distribution, and increasing $\tau$ softens the distribution. This method has been used in applications such as model calibration~\cite{temp-scaling-calib}, image restoration~\cite{temp-scaling-restoration} and image inpainting~\cite{temp-scaling-inpainting}. In this work, we use a constant temperature scale to ensure the distributional smoothness of the cross-attention maps, setting $\tau = 1.25$. Further ablation on $\tau$ is discussed in \cref{sec-further-abn}.

\begin{table*}[t]
    \centering
    \begin{tabular}{cccccccccc}
        \toprule
        $\tau$ & $T_b$ & \makecell{Source CLIP\\Score ($\uparrow$)} & \makecell{Background\\LPIPS ($\downarrow$)} & \makecell{Background\\PSNR ($\uparrow$)}  & \makecell{Background\\SSIM ($\uparrow$)} & \makecell{Structural\\Distance ($\downarrow$)} & \makecell{Target CLIP\\Score ($\uparrow$)}\\
        \midrule
        1.00 & - & 23.4216 & 0.0586 & 30.1023 & 0.8822 & 0.0728 & 25.9163\\
        1.25 & - & \textbf{23.7507} & 0.0522 & 30.4707 & 0.8849 & 0.0715 & \textbf{26.0902}\\
        1.50 & - & 24.1785 & 0.0497 & 30.7565 & 0.8863 & 0.0708 & 25.7919\\
        1.75 & - & 25.0428 & 0.0466 & 31.1206 & 0.8875 & 0.0709 & 24.9769\\
        2.00 & - & 25.4275 & \textbf{0.0409} & \textbf{31.5829} & \textbf{0.8896} & \textbf{0.0652} & 24.1544\\
        \midrule
        - & 05 & 23.5422 & 0.0562 & 30.1123 & 0.8838 & 0.0782 & 25.9403\\
        - & 10 & \textbf{23.5445} & \textbf{0.0546} & \textbf{30.3154} & \textbf{0.8847} & \textbf{0.0710} & \textbf{26.0740}\\
        - & 20 & 23.4344 & 0.0587 & 30.0937 & 0.8822 & 0.0723 & 25.8746\\
        - & 30 & 23.4494 & 0.0618 & 29.8495 & 0.8792 & 0.0757 & 25.9404\\
        - & 35 & 23.2644 & 0.0621 & 29.8123 & 0.8792 & 0.0774 & 25.8089\\
        \bottomrule
    \end{tabular}
    \caption{\label{tab:supp-ablation} \textbf{Further Ablation}: We experiment with different values of the temperature parameter ($\tau$) and bootstrap ($T_b$) parameters. From the \textit{neural} and \textit{background} metrics, we observe that the similarity between the edited and the input image increases for higher values of $\tau$ and that $T_b = 10$ is the optimal value for the bootstrap parameter.}
\end{table*}

\begin{figure*}
    \centering
\includegraphics[width=0.9\textwidth]{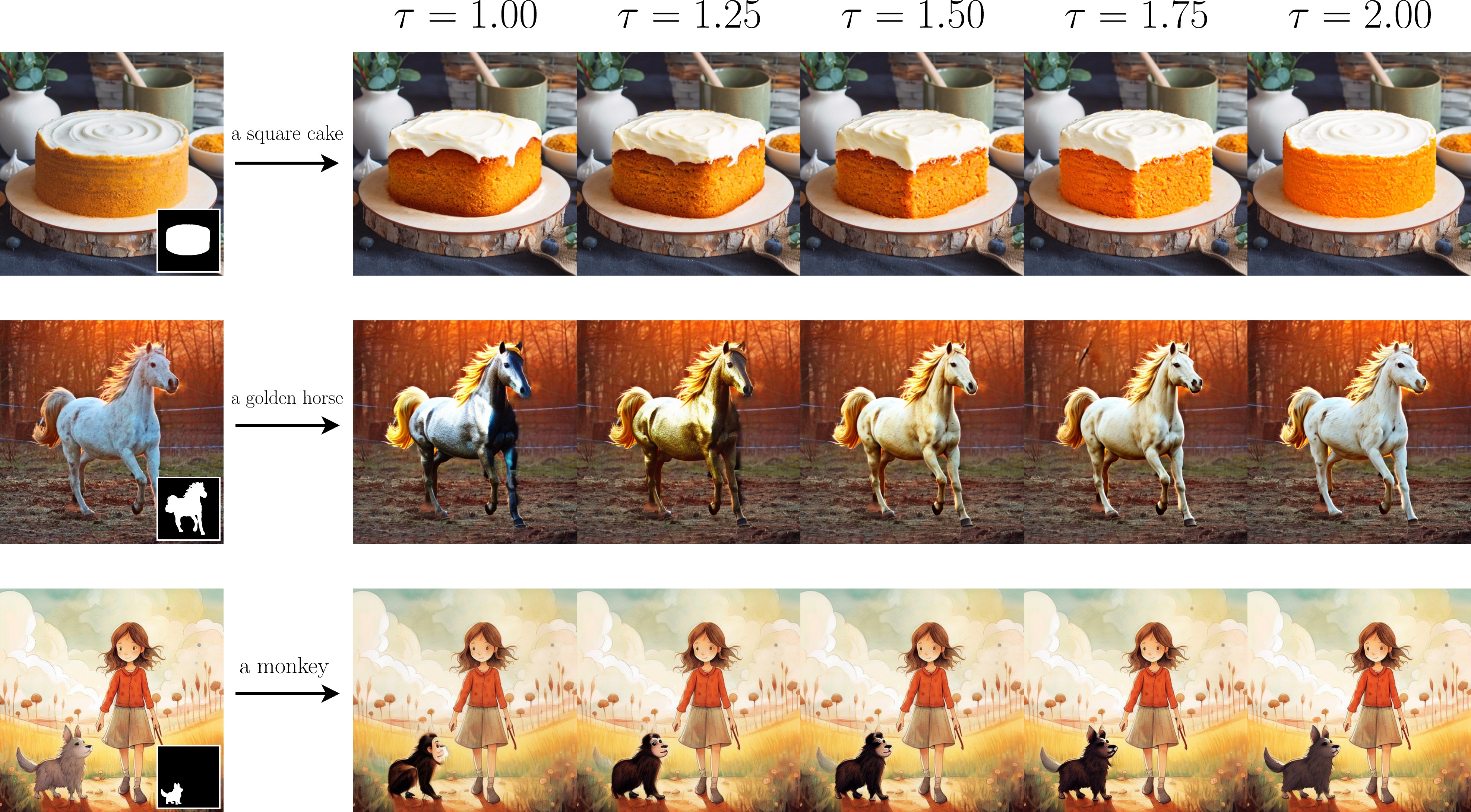}
    \caption{\textbf{Ablation on Temperature Scaling:} Impact of an increasing sequence of $\tau$'s on the edits. We observe that an increase in the value of $\tau$ results in the edited image moving towards the input image. Visually appealing edits are usually achieved at $\tau = 1.25$.}
    \label{fig:temp-scale-abn}
\end{figure*}

\section{Details on Ablation Study}
\label{sec-further-abn}

We further study the impact of varying the temperature scaling parameter $\tau$ and bootstrap $T_b$. Specifically, we experiment for $\tau \in \{1.00, 1.25, 1.50, 1.75, 2.0\}$ and $T_b \in \{5, 10, 20, 30, 35\}$ and report the results in \cref{tab:supp-ablation}.

\subsection{Temperature Scaling}

The results for variation in $\tau$ are summarized in table \cref{tab:supp-ablation} and has been depicted visually in \cref{fig:temp-scale-abn}. We observe that the edited image tends to go towards the source image with an increase in $\tau$, which can be attributed to over-smoothing the distribution. This is also indicated by the \textit{neural} metrics in \cref{tab:supp-ablation}, where an increase in $\tau$ results in increasing source CS and a decreasing target CS. This is further exemplified by the \textit{background} metrics and Structural Distance, which are the best for $\tau = 2.00$. In this work, we set $\tau = 1.25$ as mentioned in Sec. {\color{red}3.4} of the main paper. This choice of $\tau$ resulted in visually pleasing edits and we observed semantically coherent outputs for $\tau \in [1, 1.5]$.

\subsection{Bootstrap}

Upon analyzing the findings presented in \cref{tab:supp-ablation}, we opt for $T_b=10$ based on the observation that the general structure and overall layout of the image is established within the first 10 denoising steps. Subsequently, the diffusion model manifests the finer details of the image, in accordance with \cite{multidiff}. We also observe using a higher value of bootstrap aids in \textit{addition}-based edits.

\subsection{Inversion}

As mentioned in Sec. {\color{red}3.1} of the main paper, \textit{inversion} helps initiate the editing procedure and ensures a coherent and controlled edit. To understand the impact of \textit{inversion}, we compare two different initializations for the \textit{edit} process (refer Sec. {\color{red}3.2} of the main paper), namely \textbf{(1)} $x_T = x_{inv}$ and \textbf{(2)} $x_T = \zeta$. Here, $\zeta \in \mathbb{R}^{64 \times 64 \times 4}$ denotes a random latent with elements sampled from $\mathcal{N}(0,1)$. Specifically, we choose to showcase this impact on \textit{style transfer} based edits.

\begin{figure*}[t]
    \centering
\includegraphics[width=\textwidth]{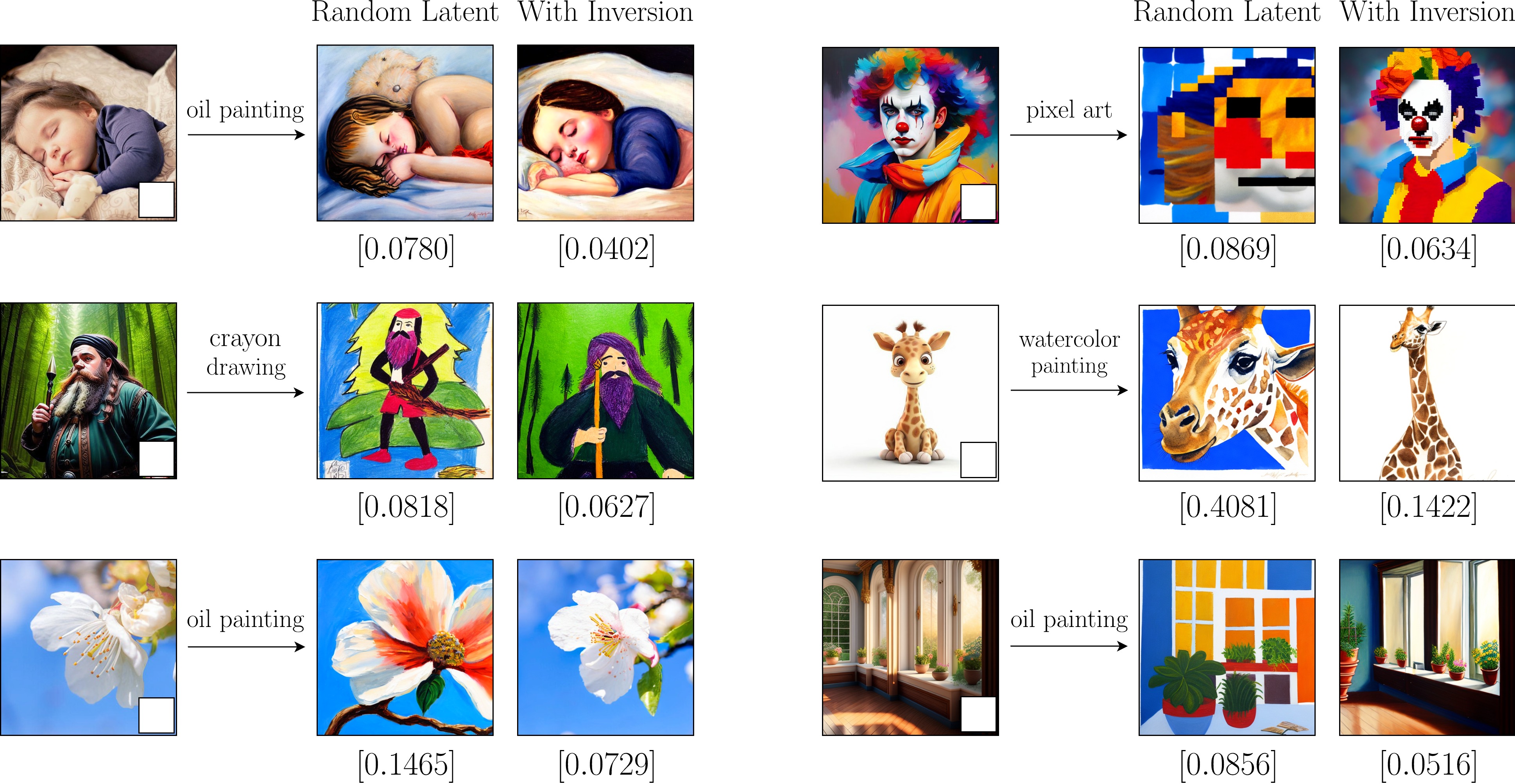}
    \caption{\textbf{Ablation on Inversion:} We study the impact of editing with a \textit{random latent} compared to initiating the editing process via \textit{inversion}. The outputs from \lomo for both cases are captioned with the [Structural Distance ($\downarrow$)]. We observe that the structural similarity is preserved when using \textit{inversion} instead of a \textit{random latent} to initiate the editing process.}
    \label{fig:inversion-supp}
\end{figure*}

From \cref{fig:inversion-supp}, we observe that the images \textit{with inversion} are structurally much closer to the input image compared to the ones generated using a \textit{random latent}, which is also indicated by the Structural Distance metric. In most cases, although using a random latent generates a faithful edit to the given prompt, it changes the content of the image, resulting in undesirable outputs. Therefore, using inversion is crucial for faithful image editing.

\subsection{Inference Time}

In a multi-object scenario, \lomo separates itself by executing all edits in a single pass, resulting in substantial time savings compared to iterative methods. This is highlighted in \cref{tab:inftime}, where our approach proves particularly advantageous in scenarios involving multiple objects, demonstrating a notable decrease in edit time. Unlike other methods that run iteratively to generate multi-object edits, \lomo's streamlined approach minimizes the need for repeated computations, enhancing overall efficiency. The gains in edit time underscore \lomo's practical applicability in real-world editing tasks, showcasing its potential to streamline and expedite complex multi-object editing processes. 

\begin{table}[h]
    \centering
    \adjustbox{max width=\linewidth}{
    \begin{tabular}{lccccc}
    \toprule
    Method & \multicolumn{5}{c}{Inference Time for $N$ masks (sec)}\\
    \cmidrule(lr){2-6}
    & 1 & 2 & 3 & 5 & 7\\
    \midrule
    GLIDE   & 22.10	& 41.10 & 63.76 & 106.99 & 153.11 \\
    DiffEdit& 22.25	& 41.30 & 65.91 & 110.85 & 152.60 \\
    BLD     & 27.20	& 51.60	& 80.40	& 135.24 & 185.37 \\
    Iterative & 23.85 & 44.67 & 70.02 & 117.69 & 163.69 \\
    \midrule
    \lomo    & 23.19 {\footnotesize\bf\color{Green} (2.8)} & 31.3 {\footnotesize\bf\color{Green} (29.9)} & 39.35 {\footnotesize\bf\color{Green} (43.8)} & 55.47 {\footnotesize\bf\color{Green} (52.9)} & 76.15 {\footnotesize\bf\color{Green} (53.5)} \\ 
    \bottomrule
    \end{tabular}
    }
    \caption{\label{tab:inftime} In a multi-object setting, we report the inference time of all the methods for varying number of masks $N$. \textit{Iterative} denotes the average runtime of GLIDE, DiffEdit and BLD. We report the percentage improvement by \lomo over \textit{Iterative} (in {\color{Green}green})}
\end{table}

\section{Experimental Protocol}

\subsection{Datasets}

To facilitate a comparison between various baselines on \textit{single-object} edits, we employ a modified subset of the PIE-Bench~\cite{direct-inversion} dataset supplemented with images from AFHQ~\cite{animal-faces}, COCO~\cite{coco}, and Imagen~\cite{imagen}. Overall, the benchmark consists of \textbf{300} images, covering editing types such as changing objects, adding objects, changing object content, changing object color, changing object material, changing background, and changing image style. Sample images for each edit type are shown in \cref{fig:single-dataset-supp}.

The newly proposed \textit{multi-object} editing benchmark \proposedDataset consists of \textbf{64} images, covering various editing types with each image featuring 2 to 7 masks, paired with corresponding text prompts. The masks for the images in \proposedDataset and the supplemental images in the \textit{single-object} dataset are generated using SAM~\cite{sam}. In practice, the user is required to provide a bounding box around the object via a GUI interface, which then automatically saves the segmented mask. Sample images from \proposedDataset are depicted in \cref{fig:multi-dataset-supp}. The images are also supplemented with various text-based annotations used by different baselines (refer \cref{tab:anno}) via a \texttt{JSON} file, including 
\begin{itemize}
    \item \textbf{Target Image Prompt (TIP)}: A complex prompt describing the complete image after the edit.
    \item \textbf{Source Mask Prompt (SMP)}: A simple text prompt describing the object inside the masked region of the input image.
    \item \textbf{Target Mask Prompt (TMP)}: A simple text prompt that describes the edited object inside the masked region.
    \item \textbf{Edit Instruction (EIn)}: Edit instruction for I-P2P~\cite{instruct-p2p}.
\end{itemize}

\begin{table}[t]
\adjustbox{max width=\linewidth}{
\begin{tabular}{@{}lcccccc@{}}
\toprule
Method & Image & Mask & \makecell{TIP} & \makecell{SMP} & \makecell{TMP} & \makecell{EIn}\\
\midrule
SDEdit \hfill \cite{sdedit}  & \cmark & \xmark & \cmark & \xmark & \xmark  & \xmark \\
I-P2P \hfill \cite{instruct-p2p}  & \cmark & \xmark & \xmark & \xmark & \xmark  & \cmark \\
DiffEdit \hfill \cite{diffedit} & \cmark & \cmark & \xmark & \cmark & \cmark  & \xmark \\
GLIDE \hfill \cite{glide} & \cmark & \cmark & \xmark & \xmark & \cmark  & \xmark \\
BLD \hfill \cite{bld} & \cmark & \cmark & \xmark & \xmark & \cmark  & \xmark \\
\midrule
\lomo    & \cmark & \cmark & \xmark & \xmark & \cmark  & \xmark \\ 
\bottomrule
\end{tabular}
}
\caption{\label{tab:anno} Annotations required by various baseline methods included in the modified \textit{single-object} dataset and \proposedDataset.}
\end{table}

\subsection{Baselines}

We use the official implementation for all baseline methods using PyTorch, except for DiffEdit as the code has not been made public. SDEdit uses the target prompt for text-guided image editing and does not require any other input. DiffEdit by construction uses the DDIM solver, but the unofficial implementation uses DPM solver~\cite{dpmsolver} for better sample efficiency. The method also generates noisy masks based on the source and target mask prompts, thus we choose to use the masks in the dataset (as mentioned in Sec. {\color{red}4.2} of the main paper).

I-P2P requires an edit instruction along with the image and does not need any other inputs. For example, the edit instruction for the first image in \cref{fig:single-edit-supp} would look like: \textit{``change the shape of the cake to a square"}. It is also important to note that although all other methods use the pre-trained Stable Diffusion model directly, Instruct-P2P is trained by finetuning this model. Finally, GLIDE and BLD are similar to \lomo in that they only require the target mask prompt as additional inputs.

\subsection{Additional Results}

\begin{figure*}[t]
    \centering
\includegraphics[width=\textwidth,height=0.95\textheight,keepaspectratio]{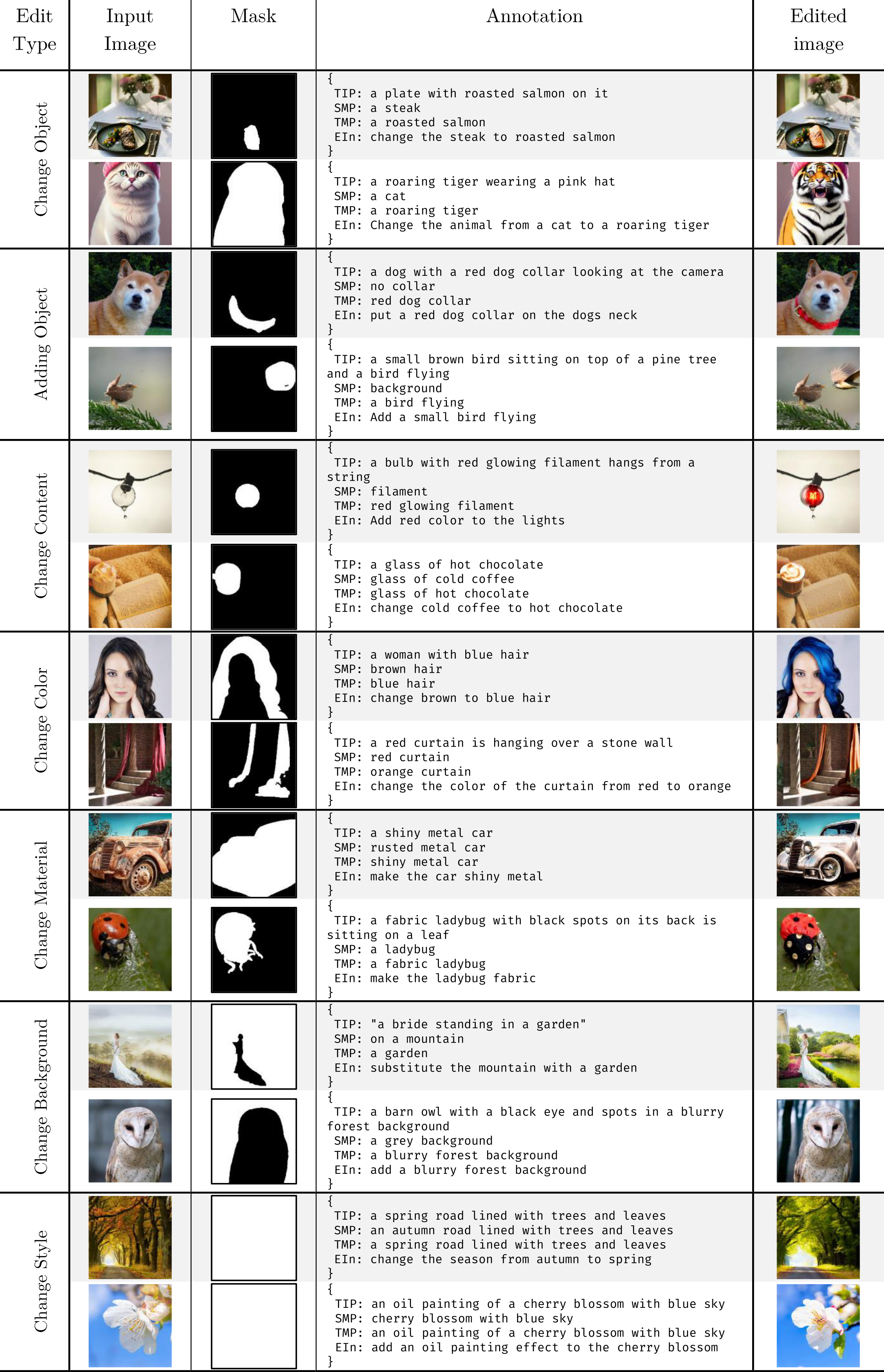}
    \caption{\textbf{Single Object Benchmark:} Examples from Single-Object dataset. The columns are \textbf{(1)} Edit type \textbf{(2)} The input image on which the editing is done, \textbf{(3)} The mask used for localizing the edit, \textbf{(4)} \texttt{JSON} annotation containing the Target Image Prompt (TIP), Source Mask Prompt (SMP), Target Mask Prompt (TMP), and the Edit Instruction (EIn), and \textbf{(5)} The edited images produced by \lomo.}
    \label{fig:single-dataset-supp}
\end{figure*}

\begin{figure*}[t]
    \centering
\includegraphics[width=0.85\textwidth]{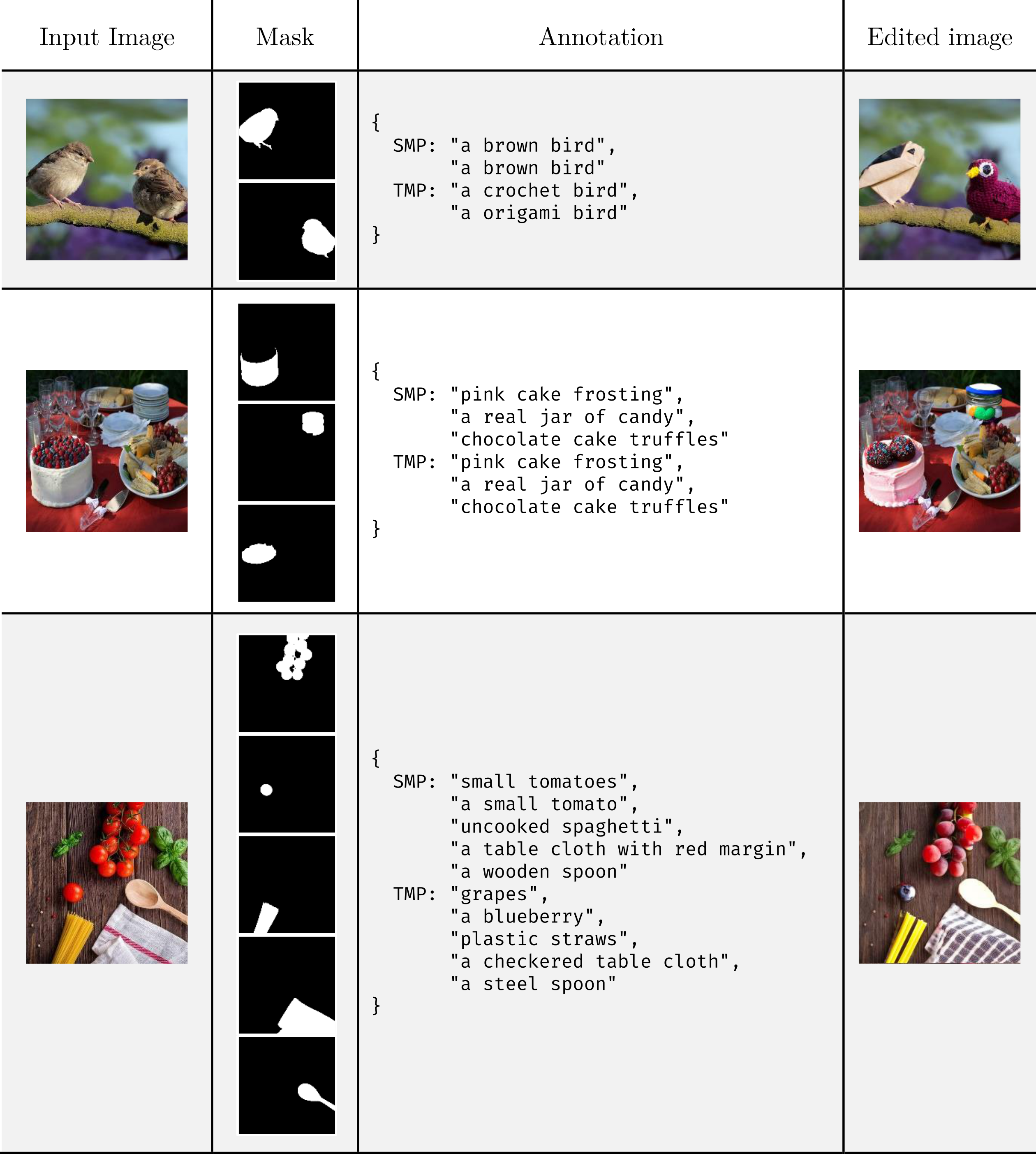}
    \caption{\textbf{\proposedDataset:} Examples from Multi-Object Dataset. The columns are \textbf{(1)} The input image on which the editing is done, \textbf{(2)} The masks used for localizing the edit, \textbf{(3)} \texttt{JSON} annotation containing the Source Mask Prompts (SMP) and Target Mask Prompts (TMP), and \textbf{(4)} The edited images produced by \lomo.}
    \label{fig:multi-dataset-supp}
\end{figure*}

We supplement the qualitative results provided in the main paper (refer Sec. {\color{red}5}) by comparing \lomo against baselines on more single-object edits, depicted in \cref{fig:single-edit-supp}. Furthermore, we showcase single-object and multi-object edits with \lomo in \cref{fig:single-dataset-supp,fig:multi-dataset-supp} for samples from \proposedDataset and the \textit{single-object} benchmark for multiple masks and various edit types, respectively.

\section{User Study}

We performed a user study using images from the \textit{single-object} dataset to assess user preferences among images edited using the various baseline methods. We had 40 participants in the age range of 23-40. The majority of them expressed a preference for the edits generated by \lomo over those from the other baseline methods. The results are summarized in \cref{fig:userstudy-supp}, and our observations from the user preference survey are as follows:

\begin{figure}[t]
    \centering
\includegraphics[width=\linewidth]{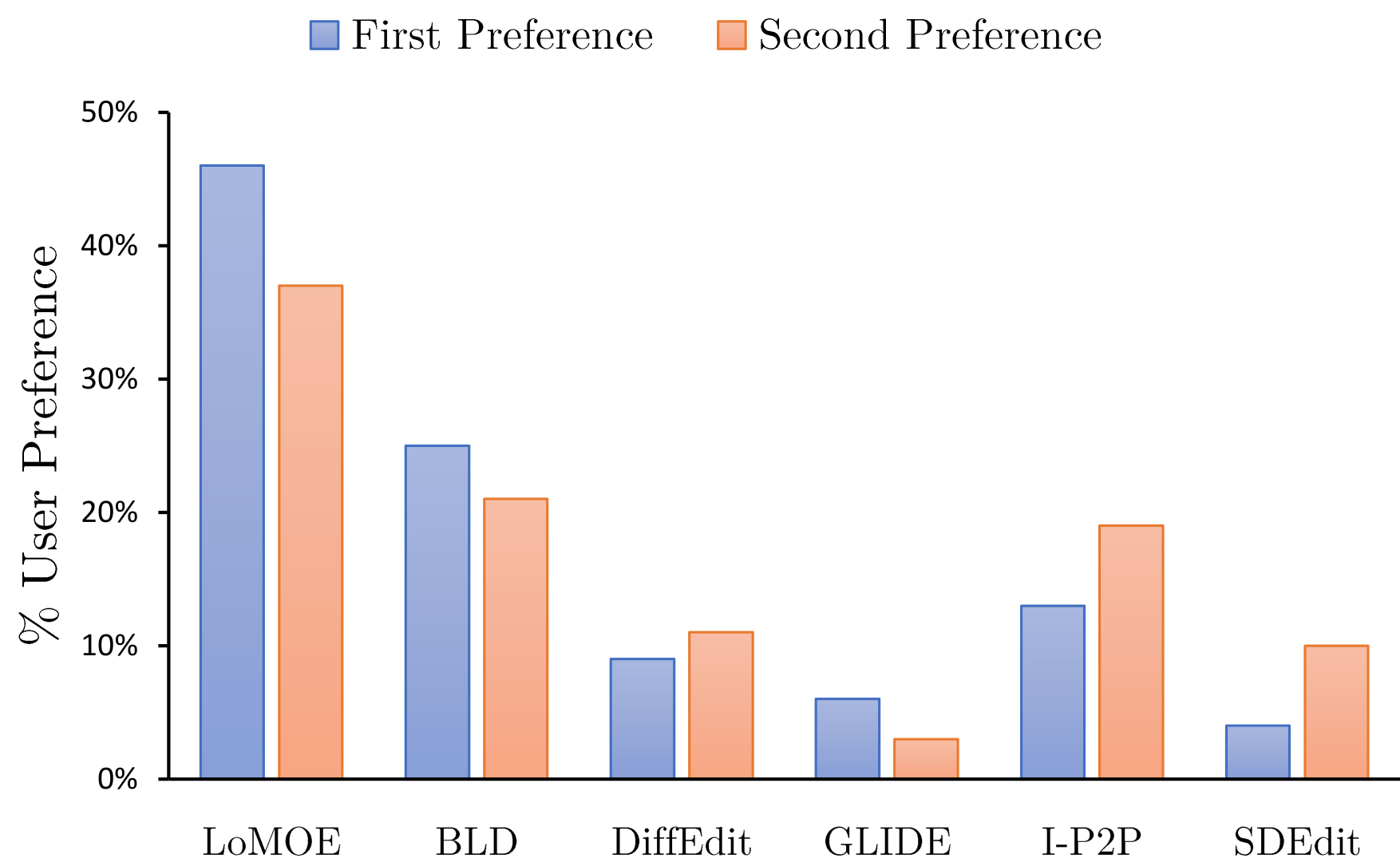}
    \caption{\textbf{User Study:} The first and second preference images for users who were shown results produced by all the above methods.}
    \label{fig:userstudy-supp}
\end{figure}

\begin{itemize}

    \item \lomo is the most preferred image editing method: It has the highest percentage of first preference (46\%) and second preference (37\%) among all the methods. This suggests that users like \lomo more than the other methods for editing images.
    
    \item Qualitatively, the users responded suggesting that \lomo could edit all images as intended, and even when the edit was not successful in a small minority of cases, \lomo didn't wrongly alter the image indicating the reliability of our method.
    
    \item BLD~\cite{bld} and I-P2P~\cite{instruct-p2p} are the second and third most preferred methods, respectively: BLD has 25\% of first preference and 21\% of second preference, while I-P2P has 13\% of first preference and 19\% of second preference. This indicates that users also appreciate BLD and I-P2P for image editing, but not as much as \lomo.
    
    \item The users were generally satisfied with BLD when it worked; however, in the cases where BLD failed, it was drastic to an extent that the edited image could no longer be used. On the other hand, I-P2P changed the background of the images; although these were unintended changes, the images produced were visually appealing. Therefore, we see I-P2P has a higher second preference than first preference.
    
    \item GLIDE~\cite{glide}, DiffEdit~\cite{diffedit}, and SDEdit~\cite{sdedit} are the least preferred methods: GLIDE has 6\% of first preference and 3\% of second preference, DiffEdit has 9\% of first preference and 11\% of second preference, and SDEdit has 4\% of first preference and 9\% of second preference. The users were not very satisfied and showed limited preference for these methods, leaning towards better alternatives.
    
    \item Qualitatively, users were dissatisfied with GLIDE as it often removed the subject to be edited and replaced it with a poor quality target. For DiffEdit and SDEdit, users noted that the images generated by both methods were very similar, except for the fact that DiffEdit preserves the unmasked region of the input image.
\end{itemize}

In conclusion, our user study provides valuable insights into user preferences amongst various baseline image editing methods. Notably, \lomo emerged as the most preferred method and users appreciated \lomo for its ability to consistently edit images as intended. They expressed dissatisfaction for other baselines due to issues such as the subject's removal, unintended changes in background, and limited visual appeal. These findings underscore the significance of user feedback in evaluating image editing methods and highlight \lomo's strong performance in meeting user expectations and generating reliable edits.

\begin{figure}[b]
    \centering
\includegraphics[width=\linewidth]{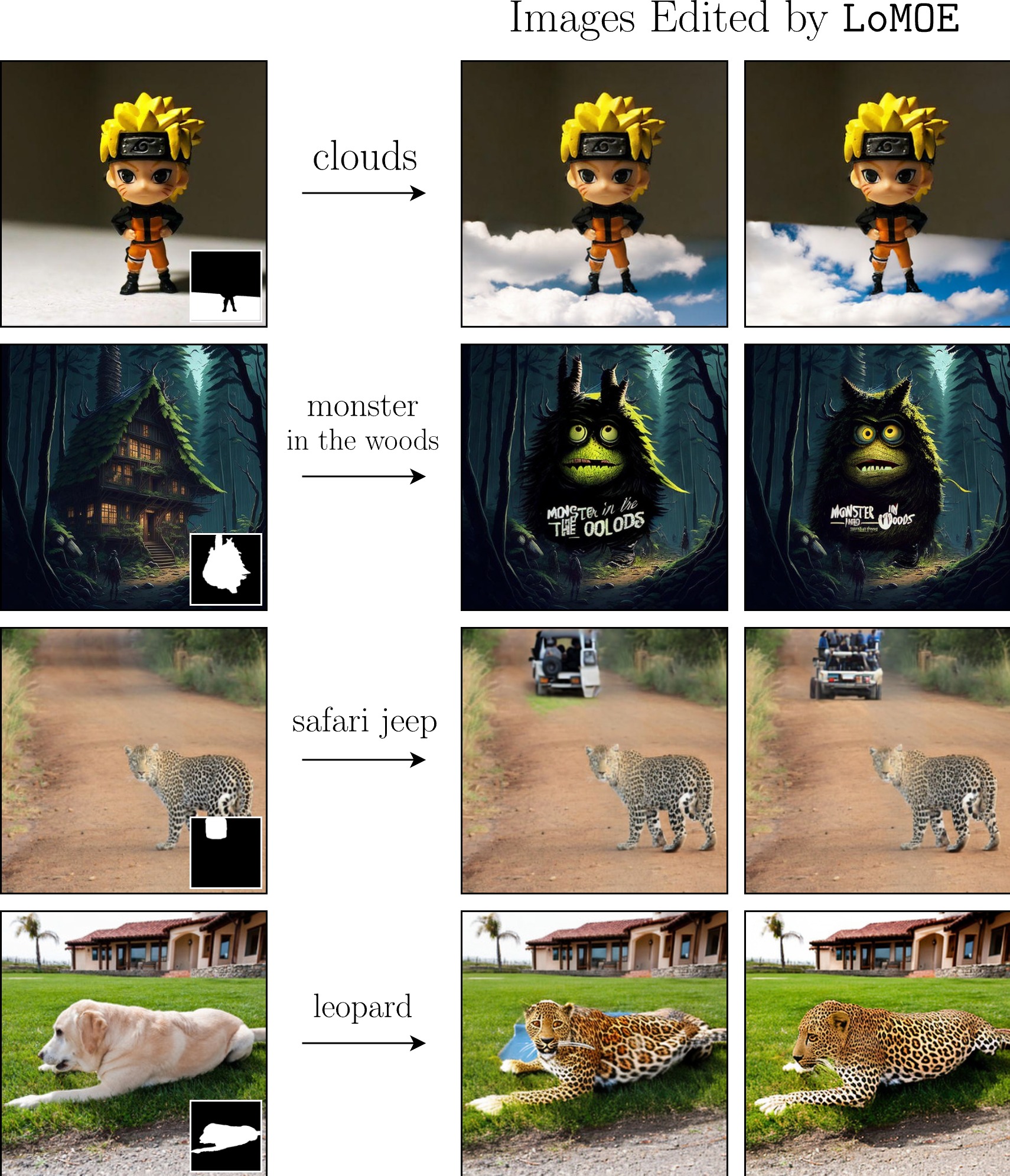}
    \caption{Illustration of \lomo's limitations revealing challenges in realism, fidelity, and object integration. Row \textbf{1} demonstrates faithful generation but the edit lacks realism. Row \textbf{2} exhibits high fidelity but includes unintended text. Rows \textbf{3} and \textbf{4} demonstrate blending inconsistencies. These limitations present promising avenues for future research.}
    \label{fig:limitations-supp}
\end{figure}

\section{Limitations}

The limitations of \lomo are depicted in Figure \ref{fig:limitations-supp}. For each example, we showcase multiple edits from the model, to fully analyze the weaknesses. In \textbf{Row 1}, although the model adheres to the prompt in adding clouds to the masked region, the edit is not very realistic, which can be attributed to the realism and faithfulness, as discussed in Section {\color{red}5.1} of the main paper. In \textbf{Row 2}, we observe although a very high fidelity edit is generated, the quote \textit{``monster in the woods"} also appears on the body of the generated object, which can be attributed to the data that the pretrained Stable Diffusion model is trained on. In \textbf{Row 3}, we observe that in the first image, the jeep doesn't blend completely with the foreground at the bottom of the mask, which is absent in the second image, and a similar trend can be observed in \textbf{Row 4}, meaning \lomo can synthesize multiple plausible results for a given prompt. 
Finally, our approach faces a limitation that is not visually evident: it cannot effectively handle object deletion or swapping within an image. This constraint opens avenues for our future research endeavors.

\section{Broader Impact}

Generative image editing models are powerful tools that can create realistic and diverse images from text or other inputs. They have many potential applications in domains such as art, entertainment, education, medicine, and security. However, they also pose significant ethical and social challenges that need to be addressed. Some of these challenges include:

\begin{itemize}
    \item The risk of generating harmful or offensive images that may violate human dignity, privacy, or rights.
    \item The possibility of manipulating or deceiving people with fake or altered images that may affect their beliefs, emotions, or behaviours.
    \item The difficulty of verifying the authenticity or provenance of images that may have legal or moral implications.
    \item The impact of replacing or reducing human creativity and agency with automated or algorithmic processes.
\end{itemize}

These challenges require careful consideration and regulation from various perspectives, such as technical, legal, ethical, and social. However, we believe that despite these drawbacks, better content creation methods will produce a net positive for society. Furthermore, we advocate for conducting such research in the public domain, emphasizing transparency and collaborative efforts to ensure responsible and beneficial outcomes for the broader community.

\clearpage


\end{document}